\documentclass[sigconf]{acmart}
\usepackage{graphicx} 
\usepackage{subfig}
\usepackage{newfloat}
\usepackage{listings}
\usepackage{multirow}

\usepackage{amsthm} 
\newtheorem{example}{Example}

\AtBeginDocument{%
  }

\copyrightyear{2026}
\acmYear{2026}
\setcopyright{cc}
\setcctype{by}
\acmConference[KDD '26]{Proceedings of the 32nd ACM SIGKDD Conference on Knowledge Discovery and Data Mining V.1}{August 09--13, 2026}{Jeju Island, Republic of Korea}
\acmBooktitle{Proceedings of the 32nd ACM SIGKDD Conference on Knowledge Discovery and Data Mining V.1 (KDD '26), August 09--13, 2026, Jeju Island, Republic of Korea}
\acmPrice{}
\acmDOI{10.1145/3770854.3780178}
\acmISBN{979-8-4007-2258-5/2026/08}

\begin{document}

\title[Improving Enzyme Prediction with Chemical Reaction Equations by Hypergraph-Enhanced \\ Knowledge Graph Embeddings]{Improving Enzyme Prediction with Chemical Reaction Equations by Hypergraph-Enhanced Knowledge Graph Embeddings}
\author{Tengwei Song}
\orcid{1234-5678-9012}
\affiliation{%
  \institution{Machine Learning Department, MBZUAI}
  \city{Abu Dhabi}
  \country{United Arab Emirates}
}
\email{songtengwei@gmail.com}

\author{Long Yin}
\affiliation{%
  \institution{Meituan}
  \city{Beijing}
  \country{China}
}
\email{yinlong02@meituan.com}

\author{Zhen Han}
\authornote{This work does not relate to Zhen Han’s position at Amazon.}
\affiliation{%
  \institution{Think Forward Lab, Amazon}
  \city{Santa Clara}
  \country{United States}}
\email{hanzhen02111@163.com}

\author{Zhiqiang Xu}
\authornote{Corresponding Author}
\affiliation{%
  \institution{Machine Learning Department, MBZUAI}
  \city{Abu Dhabi}
  \country{United Arab Emirates}
}
\email{zhiqiang.xu@mbzuai.ac.ae}

\renewcommand{\shortauthors}{Tengwei Song, Long Yin, Zhen Han, and Zhiqiang Xu}

\begin{abstract}
Predicting enzyme-substrate interactions has long been a fundamental problem in biochemistry and metabolic engineering. While existing methods could leverage databases of expert-curated enzyme-substrate pairs for models to learn from known pair interactions, the databases are often sparse, i.e., there are only limited and incomplete examples of such pairs, and also labor-intensive to maintain. This lack of sufficient training data significantly hinders the ability of traditional enzyme prediction models to generalize to unseen interactions. In this work, we try to exploit chemical reaction equations from domain-specific databases, given their easier accessibility and denser, more abundant data. However, interactions of multiple compounds, e.g., educts and products, with the same enzymes create complex relational data patterns that traditional models cannot easily capture. To tackle that, we represent chemical reaction equations as triples of (educt, enzyme, product) within a knowledge graph, such that we can take advantage of knowledge graph embedding (KGE) to infer missing enzyme-substrate pairs for graph completion. Particularly, in order to capture intricate relationships among compounds, we propose our knowledge-enhanced hypergraph model for enzyme prediction, i.e., \emph{Hyper-Enz}, which integrates a hypergraph transformer with a KGE model to learn representations of the hyper-edges that involve multiple educts and products. Also, a multi-expert paradigm is introduced to guide the learning of enzyme-substrate interactions with both the proposed model and chemical reaction equations. Experimental results show a significant improvement, with up to a 88\% relative improvement in average enzyme retrieval accuracy and 30\% improvement in pair-level prediction compared to traditional models, demonstrating the effectiveness of our approach. 
\end{abstract}




\maketitle

\newcommand{\kddavailabilityurl}{https://doi.org/10.5281/zenodo.18076811}
\ifdefined\kddavailabilityurl{}
\begingroup\small\noindent\raggedright\textbf{Resource Availability:}\\
The source code of this paper has been made publicly available at \url{\kddavailabilityurl}.
\endgroup

\section{Introduction}

Enzymes are biological catalysts that dramatically accelerate chemical reactions by lowering activation energy barriers. Identifying enzymes that catalyze specific biochemical reactions is a fundamental problem in biological systems and plays a crucial role in applications ranging from drug discovery \cite{Askr2023} to biofuel production \cite{biofuels}.

Existing machine learning approaches to enzyme prediction predominantly rely on enzyme--substrate interaction pairs curated in databases such as BRENDA \cite{BRENDA} and KEGG \cite{kegg}. While these pairs provide reliable supervision signals, they are costly to curate and thus highly incomplete, which substantially limits the generalization ability of learning-based models, especially in enzyme retrieval settings. Although recent methods, such as Boost-RS \cite{2022_boost_rs}, attempt to incorporate additional biochemical knowledge through auxiliary prediction tasks, their performance remains constrained due to the inherent sparsity of pair-level supervision and the limited exploitation of collaborative information among reactions.

In contrast, chemical reaction equations are abundantly available in domain-specific databases and explicitly describe transformations involving multiple compounds, enzymes, and products. Compared with isolated enzyme--substrate pairs, reaction equations provide richer structural context and encode collective biochemical patterns beyond individual interactions. Consequently, leveraging equation-level knowledge offers a promising direction for alleviating data sparsity in enzyme prediction.

\begin{figure}
    \centering
    \includegraphics[width=0.9\linewidth]{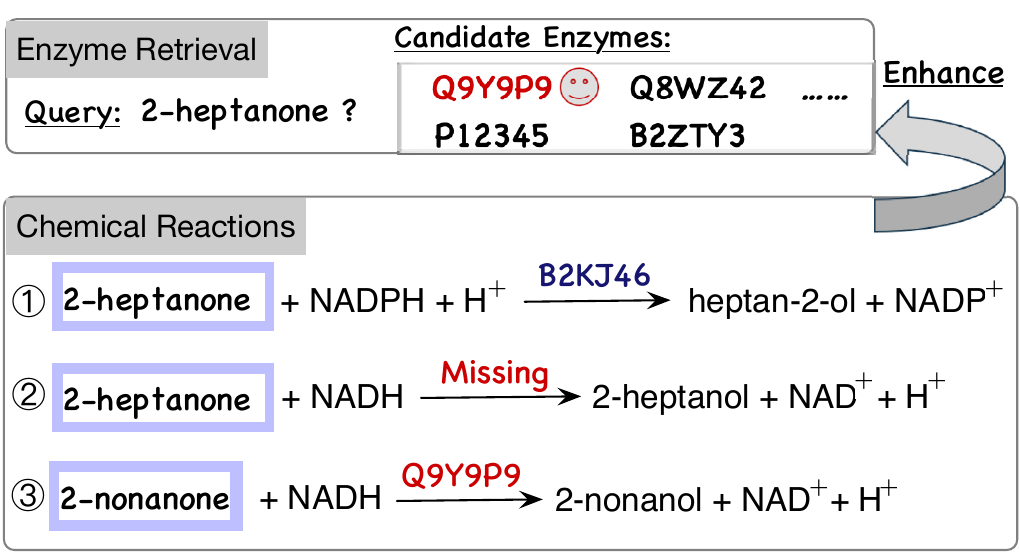}
    \caption{Enhancing enzyme retrieval using chemical reaction equations. Direct search faces two challenges: (1) the query substrate in Reaction \textcircled{1} may participate in multiple biochemical processes; (2) Reaction \textcircled{2} lacks enzyme annotation. Collaborative information from related reactions, such as Reaction \textcircled{3}, can help infer missing enzyme information.}
    \label{fig:intro}
\end{figure}

However, effectively exploiting chemical reaction equations for enzyme prediction is non-trivial. First, the overlap between enzyme--substrate pairs and reaction equations is limited. For example, among the $22{,}196$ enzyme--substrate pairs in BRENDA, only $7{,}472$ pairs can be aligned with equations containing the same compounds. Second, reaction equations often involve multiple educts and products, introducing ambiguity in identifying which compounds are directly associated with enzyme activity. As illustrated in Figure~\ref{fig:intro}, a query substrate may appear in multiple reactions catalyzed by different enzymes, resulting in substantial noise. Empirically, we observe that approximately $40\%$ of substrates can be matched to more than five compounds in reaction databases, highlighting the prevalence of high-order and ambiguous interaction patterns.

Crucially, many reaction equations in biochemical databases are incomplete, lacking precise enzyme identifiers such as PDB IDs. For instance, in the BKM-react database \cite{BKM_react}, enzymes are often annotated only by EC numbers. Rather than treating these incomplete equations as unusable noise, we view them as valuable supervision signals that encode latent relational structure across reactions. The key challenge is to effectively model the high-order compound interactions within each reaction while propagating collaborative information across different reactions to recover missing enzyme annotations.

To address this challenge, we model chemical reaction equations as a knowledge graph with triples of (educts, enzyme, products) and propose a hypergraph-enhanced knowledge graph embedding approach, termed \emph{Hyper-Enz}. Unlike conventional graph or pooling-based methods that collapse multi-compound reactions into independent pairs, Hyper-Enz explicitly captures higher-order interactions by constructing a two-level hypergraph. At the first level, educts and products in each equation are represented as heterogeneous hyperedges, preserving intra-reaction compound structure. At the second level, hyperedges are connected based on shared compounds, enabling the model to capture global collaborative patterns across reactions. The learned hyperedge representations are then integrated into a knowledge graph embedding framework, where enzymes act as relational mappings between educt and product representations, allowing missing enzyme information to be inferred via relation prediction.

Furthermore, we introduce a multi-expert decision strategy that integrates predictions from complete equations, incomplete equations, and enzyme--substrate pairs, enabling robust enzyme prediction under varying data availability. Extensive experiments demonstrate that Hyper-Enz significantly improves enzyme prediction accuracy at both the equation level and the enzyme--substrate pair level, underscoring the effectiveness of incorporating structured domain knowledge into machine learning models.

The main contributions of this work are summarized as follows:
\begin{itemize}
    \item We propose a hypergraph-enhanced knowledge graph embedding framework that captures high-order compound interactions within chemical reaction equations and heterogeneous collaborative relations across reactions.
    \item We leverage incomplete chemical reaction equations as informative supervision signals and introduce a multi-expert mechanism to integrate predictions from multiple knowledge sources.
    \item We construct a large-scale chemical reaction equation dataset by integrating multiple biochemical databases to support enzyme prediction.
    \item We conduct extensive experiments on both equation-level and enzyme--substrate pair-level enzyme prediction tasks, demonstrating consistent and significant performance improvements over state-of-the-art methods.
\end{itemize}

\section{Preliminaries and Notations}
We introduce some preliminaries and notations that we will use in this paper and summarize the notations in this paper in Table \ref{tab:notation}.

\begin{table}[H]
    \centering
    \caption{Notations and descriptions}
    \label{tab:notation}
    \begin{tabular}{cl}
    \toprule
    Notation & Description \\
    \midrule
       $M$  & Enzyme set \\
       $C$  & Compound set \\
       $q$ & A chemical equation \\
       $S$ & Set of educts in $q$ \\
       $P$ & Set of products in $q$\\
       $\mathcal{K}$ & Chemical equation knowledge graph\\
       $\mathcal{Q}$ & Chemical equation set \& Triples in $\mathcal{K}$\\
       $\mathcal{Q}'$ & Incomplete chemical equation set\\
       $\mathcal{S}$ & Set of $S$  \& Set of head entities in  $\mathcal{K}$ \\
       $\mathcal{P}$ & Set of $P$  \& Set of tail entities in  $\mathcal{K}$\\
       $S_u$ & Set of substrates \\
       $\mathcal{G}$ & Hypergraph \\
       $\mathcal{H}$ & Incidence matrix of $\mathcal{G}$ \\
       \bottomrule
    \end{tabular}
\end{table}

\subsection{Enzyme Prediction on Equation Level}

Let $M = \{m_1, m_2,\dots,m_{|M|}\}$ be the enzyme set, and $C = \{c_1, c_2,\dots,c_{|C|}\}$ the compound set, and 
$\mathcal{Q} =\{ q_1, q_2,\dots,q_{|\mathcal{Q}|}\}$ the set of all chemical reaction equations with enzyme information. Each equation is denoted by a triple $q=\langle S, m, P \rangle$ with information of a set of educts $S$ which is catalyzed by enzyme $m$ to produce a set of products $P$, where $S \subset C $, $P \subset C$, $m \in M$, and $S \cap P$ may be non-empty due to reversible reactions or shared co-factors. Also denote $\mathcal{S} = \{ S \mid \forall q = \langle S, m, P \rangle \in \mathcal{Q} \}$ and $\mathcal{P} = \{ P \mid \forall q = \langle S, m, P \rangle \in \mathcal{Q} \}$ as the set of educts and the set of products, respectively, from all equations.

To align the concepts in KGs, consider a KG \(\mathcal{K} \coloneq \{E, R, \mathcal{T}\}\), where \(E\), \(R\), and \(\mathcal{T}\) represent the set of all entities, set of all relations, and set of all triples, respectively. In this context, \(E = \mathcal{S} \cup \mathcal{P}\), \(R = M\), and \(\mathcal{T} = \mathcal{Q}\). Finally, let $\mathcal{Q}' =\{ q'_1, q'_2,\dots,q'_{|\mathcal{Q}'|}\}$ be the set of chemical reaction equations that lack valid enzyme information, associated educt set $\mathcal{S}'$, and associated product set $\mathcal{P}'$.

To formulate the enzyme prediction task into the relation prediction on the reaction equation level, given an equation KG $\mathcal{Q}$, for $q \in \mathcal{Q}$, we learn representations of $S$ and $P$, denoted as $\mathbf{S}$ and $\mathbf{P}$. The task is to let the model rank the confidence among all enzymes and predict top $k$ enzymes that the equation inclines to react with.

\subsection{Enzyme Prediction on Pair Level}
Let $S_u = \{s_{u_1}, s_{u_2}, \dots, s_{u_{|S_u|}}\}$ be the set of substrates, and $S_u \subset C$. Note that a substrate can appear simultaneously as the educt and product in a reaction equation, which means $S_u \cap S \cap P \neq \emptyset$.

To formulate the enzyme prediction task on the enzyme-substrate pair level, for any $s_i \in S_u $, we let the model predict top $k$ enzymes that the substrate inclines to react with. Note that in prior studies, the enzyme prediction was regarded as a binary classification problem. However, in reality, the experimental validation of enzyme-substrate interactions is known to be both time-consuming and expensive \cite{2023_nature_comm}. Thus, in our work, it is redefined as a retrieval task that seeks for the ranking of all potential enzymes, and allows for a more practical approach to predicting enzyme-substrate interactions.

\subsection{Hypergraph} A hypergraph, denoted as $\mathcal{G}=\{\mathcal{V}, \mathcal{E}\}$, consists of a vertex set $\mathcal{V}$ and a hyperedge set $\mathcal{E}$. Each hyperedge $ \epsilon\in \mathcal{E}$ connects two or more vertices $\nu \in \mathcal{V}$. The hypergraph can be depicted by an incidence matrix $\mathcal{H}$, where $\mathcal{H}_{\nu \epsilon} =1$ if $\nu \in \epsilon$, and $\mathcal{H}_{\nu \epsilon } =0$ otherwise.

\section{Proposed Hyper-Enz Method}
In this section, we detail the proposed Hyper-Enz method.
\subsection{Hyperedge Representation Learning}
\label{sec:hyperedge rl}
Unlike standard KGs, where each triple links two individual entities, our equation triples involve two compound sets. An intuitive solution would apply permutation-invariant pooling to aggregate each set, but this ignores interaction patterns within reactions. Instead, we construct a two-level hypergraph to model the higher-order structure among compounds.

\begin{figure}[tb]
    \centering
\includegraphics[width=\linewidth]{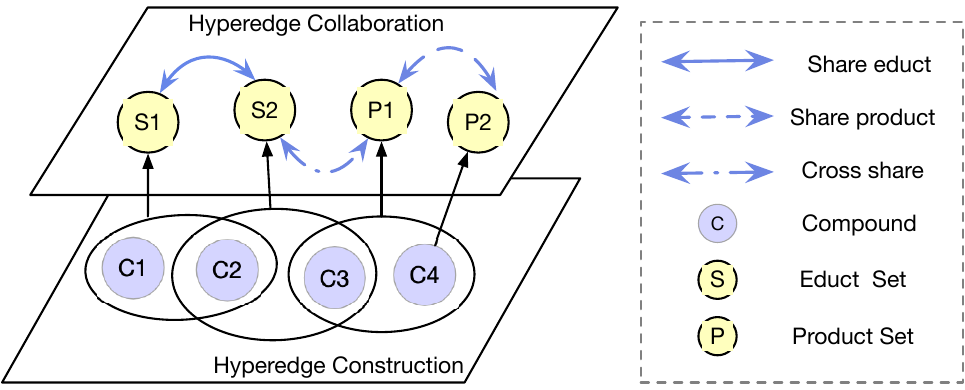}
     \caption{Hypergraph construction. Take two toy reaction equations $q_1: c_1 + c_2 \to c_3 + c_4$ and $q_2: c_2 + c_3 \to c_4$ as an example. Hyperedges are regarded as subsets of nodes in a heterogeneous graph with node types being educts and products. There are pairwise collaborative relations between hyperedges because of sharing educts, sharing products, and cross-sharing of educts and products.}
    \label{fig: hypergraph}  
\end{figure}

\begin{figure*}[htb]
    \centering    
    \includegraphics[width=\linewidth]{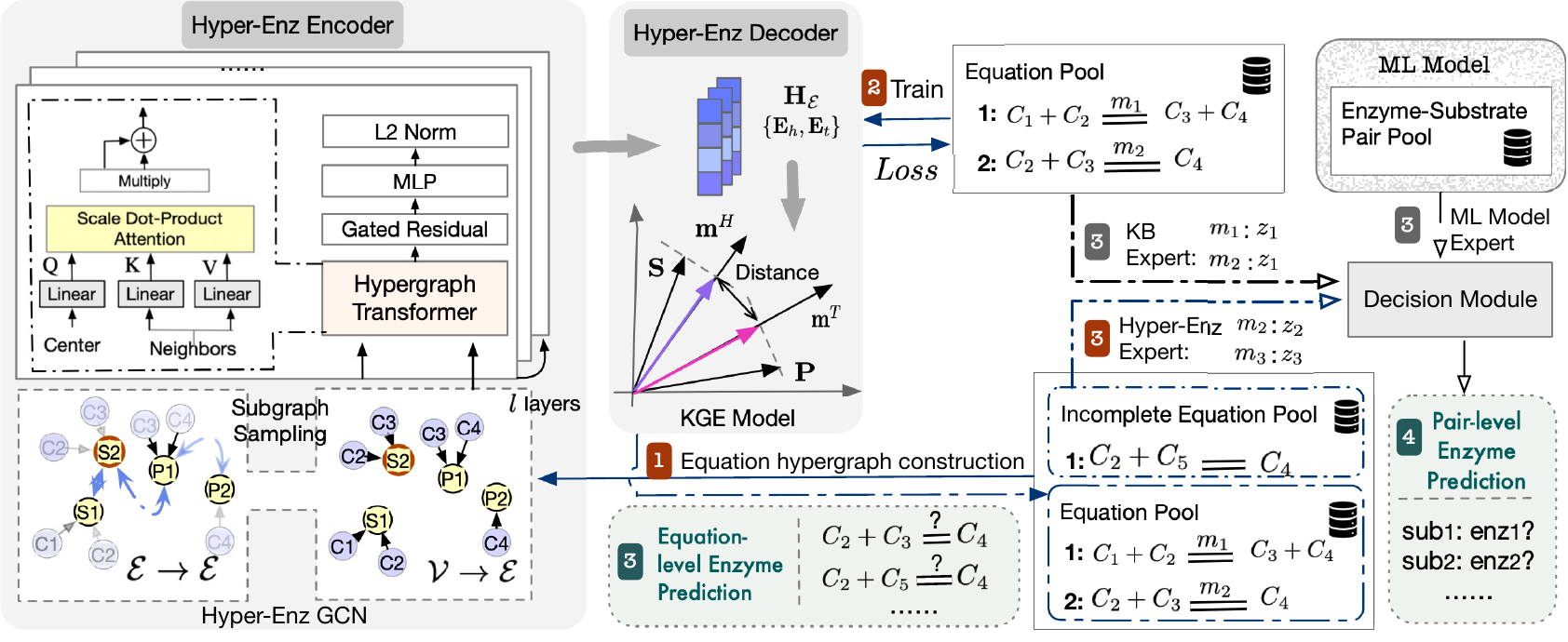}
     \caption{Training and prediction process of Hyper-Enz. \textcircled{1} We construct an equation hypergraph using both $\mathcal{Q}$ and $\mathcal{Q}'$. \textcircled{2} During training, a hypergraph transformer-based GCN encoder is used to generate representations for $\mathcal{S}$ and $\mathcal{P}$, and a KGE-based decoder to learn enzyme-aware embeddings. The encoder and decoder are optimized on $\mathcal{Q}$. \textcircled{3} In inference, we predict missing enzymes for incomplete equations in $\mathcal{Q}'$, employing a multi-expert framework that integrates outputs from a knowledge base, a pair-level ML model, and Hyper-Enz. A decision module integrates these predictions for both equation- and pair-level enzyme prediction tasks. \textcircled{4} Finally, logits from the three experts are normalized and weighted to produce a ranked top \( k \) enzyme list.}

     \label{fig: structure}
\end{figure*}

\paragraph{\bf Hypergraph construction} 
To capture the high-order collaborations of compounds beyond the equation, we adopt a hypergraph $\mathcal{G} = \{\mathcal{V},\mathcal{E}\}$ illustrated in Figure \ref{fig: hypergraph} to represent the educt set and product set in a equation, each as a hyper-edge, i.e., $\mathcal{V} \equiv C, \mathcal{E} \equiv \mathcal{S} \cup \mathcal{P} \equiv E$. To further introduce the collaborative relations among hyper-edges, we construct a matrix of relations from $\mathcal{V}$ to $\mathcal{E}$ and that of relations from $\mathcal{E}$ to $\mathcal{E}$.

We use $\mathcal{H}^{(1)} \in \mathbb{R}^{|C|\times(|\mathcal{S}| + |\mathcal{P}|)}$ and $\mathcal{H}^{(2)} \in \mathbb{R}^{(|\mathcal{S}| + |\mathcal{P}|)\times(|\mathcal{S}| + |\mathcal{P}|)}$ to denote the incidence matrix for relations from $\mathcal{V} \to \mathcal{E}$ and the hyper-edge adjacent matrix for relations from $\mathcal{E} \to \mathcal{E}$, respectively.
$\mathcal{H}^{(1)}_{k,i} =1$ if a compound $c_k \in S_i$ or $c_k \in P_i$ and 0 otherwise, while $\mathcal{H}^{(2)}_{i,j} =1$ if two hyper-edges satisfy $S_i \cap S_j \neq \emptyset$ (sharing educt), or $P_i \cap P_j \neq \emptyset$ (sharing product), or $S_i \cap P_j \neq \emptyset$ (cross sharing) and 0 otherwise, where $S_i, S_i\in\mathcal{S}$ and $P_i, P_j\in\mathcal{P}$.
Thus, relations from $\mathcal{E} \rightarrow \mathcal{E}$ are heterogeneous, with all of the above three types of correlations (i.e., sharing) covered. 

\paragraph{\bf Neighbor sampling}
For a target hyper-edge \( S \) or \( P \), we sample a set of \( \eta_1 \) hyper-edge neighbors based on \( \mathcal{H}^{(2)} \), and then further sample a set of \( \eta_2 \) 1-hop compound neighbors for all \( \eta_1 \) neighboring hyper-edges with incidence matrix \( \mathcal{H}^{(1)} \). This yields a sub-hypergraph for each target hyperedge.

\paragraph{\bf Hypergraph transformer}
We employ a hypergraph transformer to learn the representation of hyper-edges from the sub-hypergraphs obtained by neighbor sampling as inputs. The hypergraph transformer is an efficient technique for learning higher-order interactions in a hypergraph. In our experiments, a classic and widely used hypergraph transformer structure \cite{iclr2021ALLSET,sthgcn} is used.
The message-passing and propagation process in a hypergraph transformer can be described as follows:
\begin{equation}
    \mathbf{h}_i^{(l+1)}=\underset{\forall j \in \mathcal{N}(i)}{\operatorname{Propagate}}\left\{\operatorname{Message}\left(\mathbf{h}_j^{(l)}, \mathbf{r}_{i j}\right)\right\},
\end{equation}
where indices $i$ and $j$ represent the target hyper-edge and the source hyper-edge, respectively, and $\mathbf{r}_{ij}$ is the edge vector between $i$ and $j$. The set $\mathcal{N}(i)$ includes all neighbors of hyper-edge $i$. The hidden embedding $\mathbf{h}_i^{(l+1)} \in \mathbb{R}^{n}$ corresponds to the target hyper-edge $i$ at the $(l+1)$-th layer, while $\mathbf{h}_j^{(l)} \in \mathbb{R}^{n}$ represents the hidden embedding of hyper-edge $i$'s neighbor $j$ at the $l$-th layer. $\mathbf{h}_j^{(0)} = \mathbf{x}_{v}$, where $\mathbf{x}_{v}$ is the node feature of compounds. In experiments, we directly initialize $\mathbf{x}_{v}$ as a random vector in $\mathbb{R}^n$, without the need to introduce additional chemical feature information as in prior works. We use the residual connection and the layer normalization to avoid the loss of the initial information, and then utilize
multi-layer perceptrons (MLP) with a bias term to align the dimension.

After transformations through $l$ layers, we obtain representations of hyper-edges $\mathbf{H}_{\mathcal{E}}$, i.e., embeddings of head and tail entities, $\mathbf{S}, \mathbf{P}$, in the equation KG $\mathcal{Q}$.

\subsection{\bf Hyper-Enz Training and Prediction}
With representations of entities in the equation KG $\mathcal{Q}$, we adopt the current KGE method according to relation patterns in $\mathcal{Q}$ and train our Hyper-Enz model on $\mathcal{Q} \cup \mathcal{Q}'$. Also, three enzyme-predicting experts are introduced such that the model can output the final prediction by fusing results from each expert. The overall procedure for training and predicting is described in Figure \ref{fig: structure}.

\paragraph{\bf KGE-based equation-level enzyme prediction} 
We adopt a distance-based KGE method called PairRE \cite{pairre}, which proved to be able to underpin symmetry, asymmetry, and inverse relation patterns in an equation KG. More details in Appendix \ref{sec:relation-patterns}. 

For each $q \in \mathcal{Q}$, we have embeddings of its head and tail, i.e., $\mathbf{S}, \mathbf{P}$, from the hypergraph transformer, and randomly initialize the relation (enzyme) embedding as $\mathbf{m}$. A distance-based score function for PairRE is defined by
\begin{equation}
    f_r = \left\|\boldsymbol{S} \circ \boldsymbol{m}^H-\boldsymbol{P} \circ \boldsymbol{m}^T\right\|_1, 
\end{equation}
where \( \circ \) denotes the element-wise product, and \( \boldsymbol{m}^H \) and \( \boldsymbol{m}^T \) represent the head-specific and tail-specific components of the relation embedding \( \boldsymbol{m} \), respectively.  
The corresponding self-adversarial negative sampling loss \cite{rotate} is 
\begin{equation}
    \begin{aligned}
L= & -\log \sigma\left(\gamma-f_r(\boldsymbol{S}, \boldsymbol{P})\right) \\
& -\sum_{i=1}^n p\left(S_i^{\prime}, m, P_i^{\prime}\right) \log \sigma\left(f_r\left(\boldsymbol{S}_{\boldsymbol{i}}^{\prime}, \boldsymbol{P}_{\boldsymbol{i}}^{\prime}\right)-\gamma\right),
\end{aligned}
\end{equation}
where $\gamma$ is a fixed margin and $\sigma$ is the sigmoid function. Here, $\left(S_i^{\prime}, m, P_i^{\prime}\right)$ is the $i^{t h}$ negative triple, while $p\left(S_i^{\prime}, m, P_i^{\prime}\right)$ represents the weight of this negative sample defined as follows:
$$
p\left(\left(P_i^{\prime}, m, S_i^{\prime}\right) \mid(S, m, P)\right)=\frac{\exp f_r\left(\boldsymbol{S}_i^{\prime}, \boldsymbol{P}_i^{\prime}\right)}{\sum_j \exp f_r\left(\boldsymbol{S}_j^{\prime}, \boldsymbol{P}_j^{\prime}\right)} .
$$
Finally, given the Hyper-Enz model trained on $\mathcal{Q}$, we can do the relation prediction on the missing-enzyme equation KG $\mathcal{Q}'$ to impute missing enzymes. 

\paragraph{\bf Expert-guided enzyme-substrate prediction}
To improve the prediction accuracy, we further propose an expert-guided paradigm for the enzyme prediction such that the final prediction is guided by multiple experts, an idea that is akin to the Mixture of Experts (MoE) \cite{shazeer2017} but simpler because no GateNet or end-to-end training is required.
Specifically, given the query substrate $s_u$, the final enzyme prediction output $\hat{m}$ is determined by three experts. 
\begin{itemize}
    \item \textit{KB Expert} ($\mathcal{M}_1$): Retrieves enzymes by directly querying $\mathcal{Q}$ for all equations involving $s_u$.
    \item \textit{Hyper-Enz Expert} ($\mathcal{M}_2$): For $q = \langle S, m, P \rangle$ where $s_u \in S$ or $s_u \in P$, predicts a ranked list of top-$k$ enzymes for $\langle S, ?, P \rangle$ using the hypergraph-KGE model.
    \item \textit{ML Expert} ($\mathcal{M}_3$): Adapts a trained enzyme-substrate classification model to suggest candidate enzymes for $s_u$.
\end{itemize}

Based on the matching degree of \( s_u \) in the domain knowledge database, there are scenarios as follows:
\begin{itemize}
    \item $s_u \in \mathcal{S} \cup \mathcal{P} \land s_u \in \mathcal{S}' \cup \mathcal{P}'$, which means that $s_u$ is involved in both complete and incomplete chemical equations. In this case,  $\hat{m}$ can be determined by all the three experts.
    \item $s_u \notin \mathcal{S} \cup \mathcal{P} \land s_u \in \mathcal{S}' \cup \mathcal{P}'$, which means that $s_u$ only appears in $\mathcal{Q}'$. In this case, we can utilize the Hyper-Enz expert $\mathcal{M}_2$ and the ML model expert $\mathcal{M}_3$ to assist in the prediction of $\hat{m}$.
    \item $s_u \in \mathcal{S} \cup \mathcal{P} \land s_u \notin \mathcal{S}' \cup \mathcal{P}'$, which means that $s_u$ only appears in $\mathcal{Q}$. In this case,  $\hat{m}$ is determined by the KB expert $\mathcal{M}_1$ and ML model expert $\mathcal{M}_3$.
    \item $s_u \notin \mathcal{S} \cup \mathcal{P} \land s_u \notin \mathcal{S}' \cup \mathcal{P}'$, which means that $s_u$ is not involved in the chemical-reaction domain knowledge database, and the result directly comes from the ML model expert $\mathcal{M}_3$.
\end{itemize}


\paragraph{\bf Decision module} To combine the top-k predictions and logits from three experts and obtain the final top-k predictions, we introduce a decision module. Denote \( z_{i,j} \) as the logit of the \( j \)th predicted result from the $\mathcal{M}_i$ expert, 
the decision model first uses z-score normalization to ensure the logits are on a similar scale and a weighted sum of the logit of each enzyme is computed based on the standardized logits:
\begin{equation}
    \hat{z}_{i,j} = \frac{z_{i,j} - \mu_i}{\sigma_i},\quad
    m_j = \sum_{i=1}^{3} w_i \cdot \hat{z}_{i,j},
\end{equation}
where \(\mu_i\) and \(\sigma_i\) are the mean and standard deviation of the logits for the  $\mathcal{M}_i$ expert, respectively, and \( w_i \) is the weight of logits for the $\mathcal{M}_i$ expert.
 Finally, the top-$k$ enzyme-prediction results is obtained by a softmax operation: $\hat{m}_j = \frac{e^{m_j}}{\sum_{l=1}^{K} e^{m_l}}$.

\section{Experiments}
In this section, we empirically study the performance of Hyper-Enz on equation-level and pair-level enzyme prediction. All codes are implemented
based on the CentOS Linux 7, 4 $\times$ 42G A100 in PyTorch \cite{pytorch}.

\subsection{Equation-level Enzyme Prediction}
We introduce a novel task: predicting enzymes from educts and products in chemical reactions. This task validates the efficacy of Hyper-Enz in the relation prediction and holds practical significance because data on enzyme-substrate pairs is scarce in practice, contrasting with abundant chemical equation data. When specialized pairs lack required substrates, leveraging chemical equations and models like Hyper-Enz for relation prediction proves a valuable strategy.

\subsubsection{\bf Experimental Setup}
\paragraph{\bf Dataset.} We construct a reaction equation dataset, named EQ50k from chemical reaction databases BRENDA \cite{BRENDA} and PubChem \cite{pubchem2023}. It contains $38,861$ complete equations $\mathcal{Q}$, with $15,772$ compounds and $17,401$ enzymes. Additionally, EQ50k contains $42,714$ incomplete equations $\mathcal{Q}'$ missing enzymes, where compounds have appeared in $\mathcal{Q}$. To evaluate the models, we split $\mathcal{Q}$ into training, validation, and test sets at an $8:1:1$ ratio.
We collect our EQ50k-$\mathcal{Q}$ and EQ50k-$\mathcal{Q}'$ from the following databases:
\begin{itemize}  
    \item BRENDA \cite{BRENDA}. A widely used enzyme database with detailed information on enzyme function, EC classifications, and reaction mechanisms. Among its $82,119$ reactions, $75,483$ lack a PDB ID, which we include in EQ50k-$\mathcal{Q}'$.  
    \item BKMS-react \cite{BKM_react}. A comprehensive database integrating biochemical reactions from BRENDA, KEGG, MetaCyc, and SABIO-RK. It aligns substrates and products for consistency, representing enzymes by EC Numbers, all of which contribute to EQ50k-$\mathcal{Q}'$.  
    \item PubChem \cite{pubchem2023}. A large chemical database from NCBI, offering compound structures, properties, and biological activities. Since the other databases use compound names while enzyme-substrate pairs \cite{enz_rank2023} rely on SMILES notation, PubChem enables conversion between these representations.  
\end{itemize}

\paragraph{\bf Baseline models.} We adopt typical KGE and GNN methods as our baseline model on equation-level enzyme prediction. 
The KGE methods include TransE \cite{transe}, RotatE \cite{rotate}, and PairRE \cite{pairre}, ComplEx \cite{CompLex} and Rot-Pro \cite{rot-pro}. The GNN methods include LighntGCN \cite{caisimple} and HL-GNN \cite{zhang2024heuristic}.To adapt these models to our task, we use the mean representation of the compound sets corresponding to educts and products, treating them as the head and tail entities.

\paragraph{\bf Parameter settings.}
The hyperparameters in training Hyper-Enz are as follows: learning rate $10^{-3}$ and batch size $512$. The embedding dimension of all compounds and enzymes is searched from $\{ 64, 128, 256, 512\}$, hyperedge sampling sizes $\eta_1$ from $ \{5, 10, 30, 50\}$, $\eta_2$ from $\{1, 3, 5, 10\}$, negative sampling size from $\{100,200,400\}$, dropout rate from $ \{0.1, 0.3, 0.5\}$, and regularization weight from $\{0.001, 0.005, 0.01, 0.05\}$.

\subsubsection{\bf Results and Analysis}
\paragraph{\bf Overall performance.} We do relation prediction with Hyper-Enz and baseline KGE methods on EQ50k-$\mathcal{Q}$. We report results in Table \ref{tab:exp_res_with_kge} in terms of MR, MRR, and  Hit@$k$ for $k=1,3,10$, where the best results are in \textbf{bold} and the second-best are \underline{underlined}.

We consider three variants of Hyper-Enz: Hyper-Enz (Homo), Hyper-Enz (MLP), and Hyper-Enz (Random). Hyper-Enz (Homo) treats the underlying hypergraph as homogeneous by not distinguishing between educt and product hyperedges, where all three types of hyperedge-sharing relations are considered identical. Hyper-Enz (MLP) replaces the PairRE scoring function with an MLP-based predictor. Hyper-Enz (Random) uses random initialization for compound and enzyme representations instead of features derived from SMILES strings and EC numbers.

\begin{table}[htb]
    \centering
     \caption{Relation prediction results of Hyper-Enz with KGE and GNN methods}
    \label{tab:exp_res_with_kge}
     \resizebox{\linewidth}{!}{
    \begin{tabular}{lccccc}
        \toprule
         &  MR & MRR & H@1 & H@3 & H@10  \\
        \midrule
        TransE &  3326 & .194 & .099 & .236 & .400 \\
        ComplEx & 4826 & .150 & .077 & .183 &  .303 \\
        RotatE  & 4202 & .160 & .084 & .198 & .324 \\
        Rot-Pro & 2102 &.181 & .094 & .228 & .363 \\
        PairRE & 1977 & .223 & .103 & .233 & .378 \\
        \midrule
        LightGCL & - & .206 & .099& .230 & .440 \\
        HL-GNN	& - & .199 &	.127 &	.217	& .346 \\
        \midrule
    Hyper-Enz(Homo) & 2472 & .265 & .186 &.299 &.431 \\
        Hyper-Enz(MLP)& 2297 &\underline{.275 }& .184 & {.301}  & \underline{.452} \\ 
        Hyper-Enz(Random) & 2497 & .271 & \underline{.189} & \underline{.309} & .438
 \\ 
  Hyper-Enz & \textbf{1936} & \textbf{.299} & \textbf{.223} & \textbf{.326} & \textbf{.458} \\
        \bottomrule
    \end{tabular}
    }
\end{table}

Results show that Hyper-Enz outperforms the baseline models in terms of all evaluation metrics. Additionally, the significant improvement of Hyper-Enz over Hyper-Enz(Homo) indicates the effectiveness of our heterogeneous hyper-edge collaboration mechanism. It is also notable that 
Hyper-Enz(MLP) shows compatible performance with Hyper-Enz, illustrating that even the simplest embedding model can also effectively learn the relations between entities. The results on variants of Hyper-Enz(Random) ans  Hyper-Enz also show that incorporating additional structural information has a positive effect on performance. 

\paragraph{\bf Effect of hyper-edge collaboration.}
Our Hyper-Enz model incorporates three types of sharing relations: sharing educt, sharing product, and cross sharing, all of which rely on hyper-edge collaboration. To examine the effect of this collaborative information, we conduct experiments with different hyper-edge sampling sizes. As shown in Figure~\ref{fig:sampling size}, (a) reports model performance with the 2-hop neighborhood size ($\eta_2 = 5$) fixed while varying the 1-hop size ($\eta_1 \in \{5,10,30,50\}$). In (b), we fix $\eta_2 = 10$ and vary $\eta_1 \in \{1,3,5,10\}$.

\begin{figure}[h]
  \centering
  \subfloat[1-hop neighborhood size]{\includegraphics[width=0.5\linewidth]{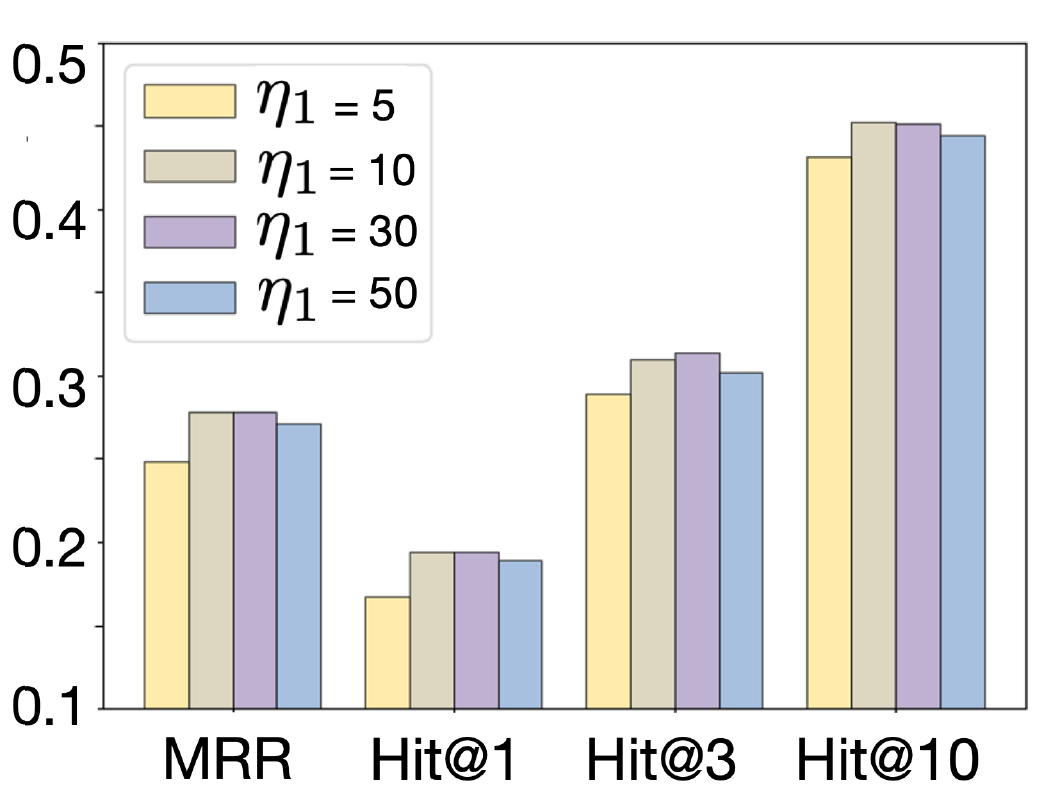}\label{fig:1-hop neighbor size}}
  \hfill
  \subfloat[2-hop neighborhood size]{\includegraphics[width=0.5\linewidth]{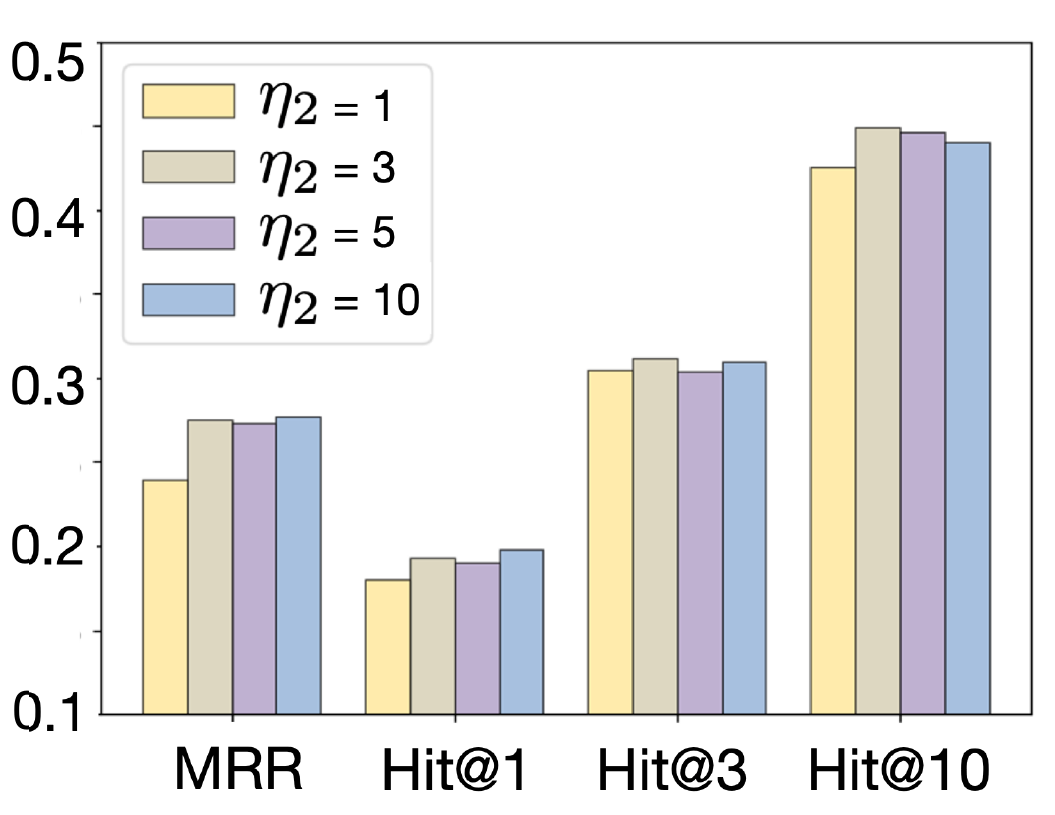}\label{fig:2-hop neighbor size}}
  \caption{Effect of neighbor sampling size on model performance.}
  \label{fig:sampling size}
\end{figure}

We can see from the result that with smaller hyper-edge sampling sizes, the performance of Hyper-Enz generally slightly decreases, highlighting the importance of sufficient collaborative information for accurate enzyme-reaction predictions. This suggests that smaller sampling sizes may not capture enough information for the model to make effective use of the hyper-edge collaboration. As the hyper-edge sampling size increases, however, the trend tends to plateau after reaching an optimal sampling size, suggesting that after all the neighbors are sampled, there is no significant contribution to further improvements.
\begin{table*}[htb]
  \centering
  \caption{Enzyme prediction results on enzyme-substrate pair datasets with different test set divisions.}
  \label{tab:res on es-pair}
  \begin{tabular}{lccccc ccccc}
    \toprule
    & 
    \multicolumn{5}{c}{ES-23k} &
    \multicolumn{5}{c}{ES-23k-M} 
    \\
    \cmidrule(lr{.75em}){2-6} 
\cmidrule(lr{.75em}){7-11} 
     & MR & MRR & Hit@1 & Hit@3 & Hit@10 
     & MR & MRR & Hit@1 & Hit@3 & Hit@10 
\\
\midrule
     BPR & - &  .094 & .071 & .127 & .228 & - & .067 &  .043 &  .100  &  .129\\
     DGCF & - & .100 & .074 & .136 & .247 & - &  .086 &    .059  &  .122  &  .168 \\
     NGCF & - & .106 & .080  & .139 & .244 & - &  .069 & .040 &   .106 &  .218 \\
     Boost-RS & 970 &  .021 &  .007 & .144 & .209 & 1240 & .008 & .000  & .075 & .116 \\
     Boost-RS(CC) & 946 & .017 & .002 & .142 & .207 & 1184 & .009 & .002 & .079 & .114 \\
     EnzRank & \underline{88} & \underline{.193}  &  \underline{.118} & \underline{.189} & \underline{.330} & \underline{212} & \underline{.165} & \underline{.108} & \underline{.153} & \underline{.269} \\
    MEI\textsubscript{GCN} & 614 & .032 & .018 & .035 & .068 & 621 & .030 & .016 & .032 & .063 \\
     FusionESP & 702 & .038 & .025 & .041 & .077 & 685 & .034 & .022 & .038 & .071 \\
\midrule
     Hyper-Enz & \textbf{83} & \textbf{.209} & \textbf{.130} & \textbf{.205} & \textbf{.368} & \textbf{184} & \textbf{.215} & \textbf{.144}  & \textbf{.222} & \textbf{.356} \\
\bottomrule
\end{tabular}
\end{table*}

\subsection{Enzyme-Substrate Prediction}
In this task, we directly apply the multi-expert mechanism on top of Hyper-Enz for the prediction of enzyme-compound pairs, without the need for retraining the model. Additionally, we set the ML Model Expert as EnzRank, a state-of-the-art model for enzyme-compound prediction
.
\subsubsection{\bf Experimental Setup}
\paragraph{\bf Dataset.} We follow the enzyme-substrate pair dataset published by EnzRank \cite{enz_rank2023}\footnote{\url{https://github.com/maranasgroup/EnzRank}}. It is constructed based on BRENDA database, with $22,196$ enzyme-substrate pair, $12,196$ enzymes, and 10,170 substrates. We name it ES-23k for easier reference later.

We matched the test set with both our EQ50k-$\mathcal{Q}$ and EQ50k-$\mathcal{Q}'$. Among the compounds in EQ50k-$\mathcal{Q}$ and EQ50k-$\mathcal{Q}'$, we identified that $612$ pairs out of $2,220$ testing pairs can match both sets. Additionally, $758$ pairs can only match $\mathcal{Q}$, and $998$ pairs can only match $\mathcal{Q}'$.
To better validate the efficiency of our model, we re-divided the training, validation, and test sets while maintaining the same ratio. Enzyme-compound pairs that align with domain knowledge were placed into the test set. We name it as ES-23k-M. The test set of ES-23k-M is also sparser than ES-23k.
We conduct experiments on both ES-23k and ES-23k-M.

\paragraph{\bf Baseline models.} 
We compare Hyper-Enz with classic recommendation methods, including Bayesian Personalized Ranking (BPR) \cite{bpr,DMF}, Neural Matrix Factorization (NMF) \cite{NCF}, Neural Graph Collaborative Filtering (NGCF) \cite{NGCF}, and Disentangled Graph Collaborative Filtering (DGCF) \cite{DGCF}. We also include enzyme-substrate–specific models such as EnzRank \cite{enz_rank2023}, Boost-RS \cite{2022_boost_rs} and its variant Boost-RS(CC) (which uses only compound–compound bio-transformations as an auxiliary task), MEI\textsubscript{GCN} \cite{mei}, and FusionESP \cite{FusionESP}.
Since all baseline models are trained and evaluated as binary classification tasks, during training, we apply random negative sampling, enabling them to return multiple prediction rankings more effectively during the prediction phase.

\paragraph{\bf Parameter settings.} During the enzyme-substrate prediction stage, there are three main hyper-parameters that influence the prediction results: $w_1$ as the weight of the KB Expert, $w_2$ as the weight of the Hyper-Enz Expert, and $w_3$ as the weight of the ML model expert. These weights are optimized by searching across several predefined combinations, specifically $(w_1,w_2,w_3) \in 
 \{(0.1,0.1,0.8)$, 
 $(0.1,0.3,0.6)$, 
 $(0.1,0.7,0.2)$, $(0.3,0.1,0.6)$, 
 $(0.4, 0.1, 0.5)$, 
 $(0.7,0.1,0.2)\}$.

\paragraph{\bf Hypergraph construction for testing.}
 For the equation-level enzyme prediction task, given a test triple  $(S, e?, P)$,
 there might be a new hyper-edge $\epsilon_s$ and  $\epsilon_p$ which has not appeared previously. To get the representation of $\epsilon_s$ and $\epsilon_p$, we use the existing collaboration from the training data to calculate the sharing information via the product incidence matrices of training and testing. The representation of the new hyperedge is obtained through forward propagation, as described in Section \ref{sec:hyperedge rl}.
The message-passing process during inference is illustrated in Figure \ref{fig:hypergraph_testing}.
\begin{figure}[h]
    \centering
    \includegraphics[width=0.7\linewidth]{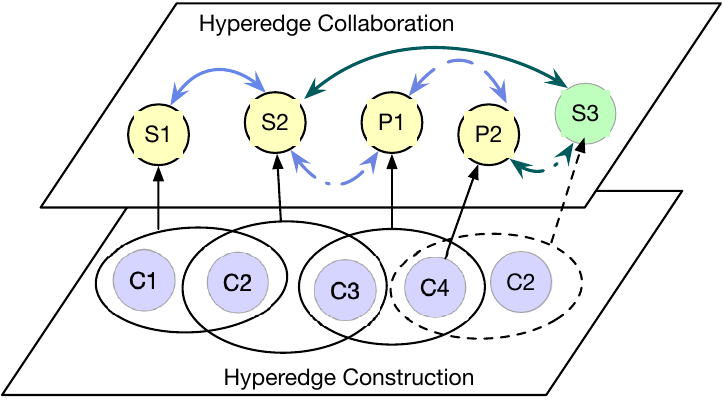}
    \caption{Hypergraph construction during inference. Suppose that the testing educt set $S=\{c_2,c_4\}$.}
    \label{fig:hypergraph_testing}
\end{figure}

\paragraph{\bf Evaluation protocol.} 
We evaluate the models using metrics: mean rank (MR), mean reciprocal rank (MRR), and top-$k$ Hit Ratio, i.e., Hit@$k$. 
For each triple $(\mathbf{S}, e, \mathbf{P})$ in the test set, we generate corrupted triples for the relation prediction task by replacing either $e$ with every other enzyme in the dataset. In line with prior research \citep{transe}, all models are evaluated in a \emph{filtered} set, meaning corrupted triples found in the training, validation, or test sets are excluded from the ranking process. The valid triple and the filtered corrupted triples are then ranked in ascending order based on their prediction scores. A lower MR, higher MRR, and higher Hit@$k$ signify better performance.

\subsubsection{\bf Results and Analysis}
\paragraph{\bf Overall performance.} We conduct enzyme prediction at the pair level, employing the same metrics as the equation-level prediction. The enzyme prediction results of Hyper-Enz and baseline models on ES-23k and ES-23k-M are shown in Table \ref{tab:res on es-pair}, where the best results are in \textbf{bold} and the second-best are \underline{underlined}. We can see that Hyper-Enz consistently outperforms the baseline models across all metrics, demonstrating the effectiveness of incorporating hypergraph structures and expert knowledge into the prediction process. 
In particular, the enhancement observed on ES-23k-M is more pronounced than that on ES-23k, which aligns with our expectations as ES-23k-M can leverage a wider range of domain knowledge during testing. The investigation into the gains provided by various prediction experts will be further explored in subsequent sections.

\paragraph{\bf Ablation study.} 
We conduct an ablation study on the subset of ES-23k-M by removing or modifying certain predicting expert of the Hyper-Enz to see their influence on the results. The subset contains $2,088$ pairs which can match both $\mathcal{Q}$ and $\mathcal{Q}'$. 
Specifically, we consider that the superiority of Hyper-Enz can be seen as the result of the joint effect of the following: 1) hypergraph transformer; 2) KGE method; 3) KB Expert; 4) Hyper-Enz Expert. We replace the hypergraph modeling with a mean pooling, replace the KGE method with MLP, and remove the KB Expert and Hyper-Enz Expert, respectively. The ablation study is shown in Table \ref{tab:ablation_exp}.

\begin{table}[htb]
    \centering
     \caption{Ablation study on Hyper-Enz}
    \label{tab:ablation_exp}
     \resizebox{\linewidth}{!}{
    \begin{tabular}{lcccc}
        \toprule
         & MRR & H@1 & H@3 & H@10  \\
        \midrule
        \textbf{Full Model} & \textbf{.215} & \textbf{.144}  & \textbf{.222} & \textbf{.356} \\
        w/o KGE & .174 & .118 & .170 & .276\\
        w/o Hypergraph &      .165 & .108 & .153 & .269 \\
        w/o KB Expert &      .168 & .114 & .166 & .264  \\
        w/o Hyper-Enz Expert & .170 & .117 & .153 & .284 \\
        \bottomrule
    \end{tabular}
    }
\end{table}

As can be observed, each component impacts the full
model performance to a certain extent, which indicates the necessity of Hyper-Enz to incorporate these mechanisms. Furthermore, it is worth noting that although Hyper-Enz(MLP) gains competitive results on the equation-level prediction in previous experiments, its performance in the pair-level prediction was not equally strong. One possible reason for this discrepancy is the limited size of the test set in the equation dataset for which Hyper-Enz(MLP) suffices to fit adequately. However, when predicting on a large-scale and incomplete equation dataset, the KGE method is more advantageous.

\paragraph{\bf Hyper-parameter sensitivity study.}
We investigate the effect of the weight of three experts. Specifically, we select the weight of Hyper-Enz $w_2$ from $\{0.1, 0.3, 0.5, 0.7, 0.9\}$ and let $w_1 = w_3 =\frac{1}{2}(1-w_2)$. Similarly, we select the weight of KB Expert $w_1$ from $\{0.1, 0.3, 0.5, 0.7, 0.9\}$ and let $w_2 = w_3 =\frac{1}{2}(1-w_1)$. The impact of weights from different experts on the results is illustrated by the line graph in Figure \ref{fig:weight of experts}. 

\begin{figure}[h]
  \centering

  \subfloat[Effect of $w_1$]{\includegraphics[width=0.45\linewidth]{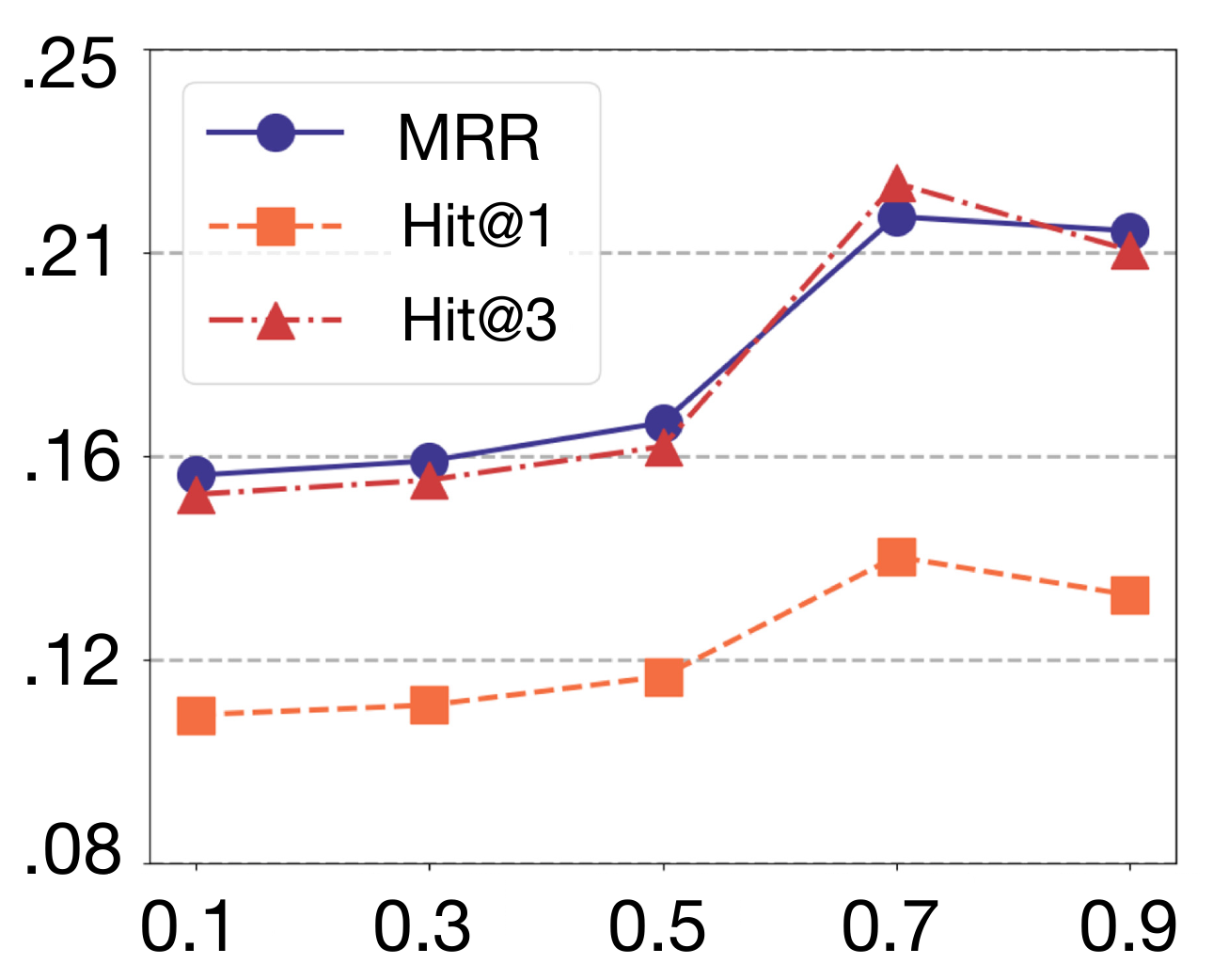}}\label{fig:w_1 plot}
  \hfill
  \subfloat[Effect of $w_2$]{\includegraphics[width=0.45\linewidth]{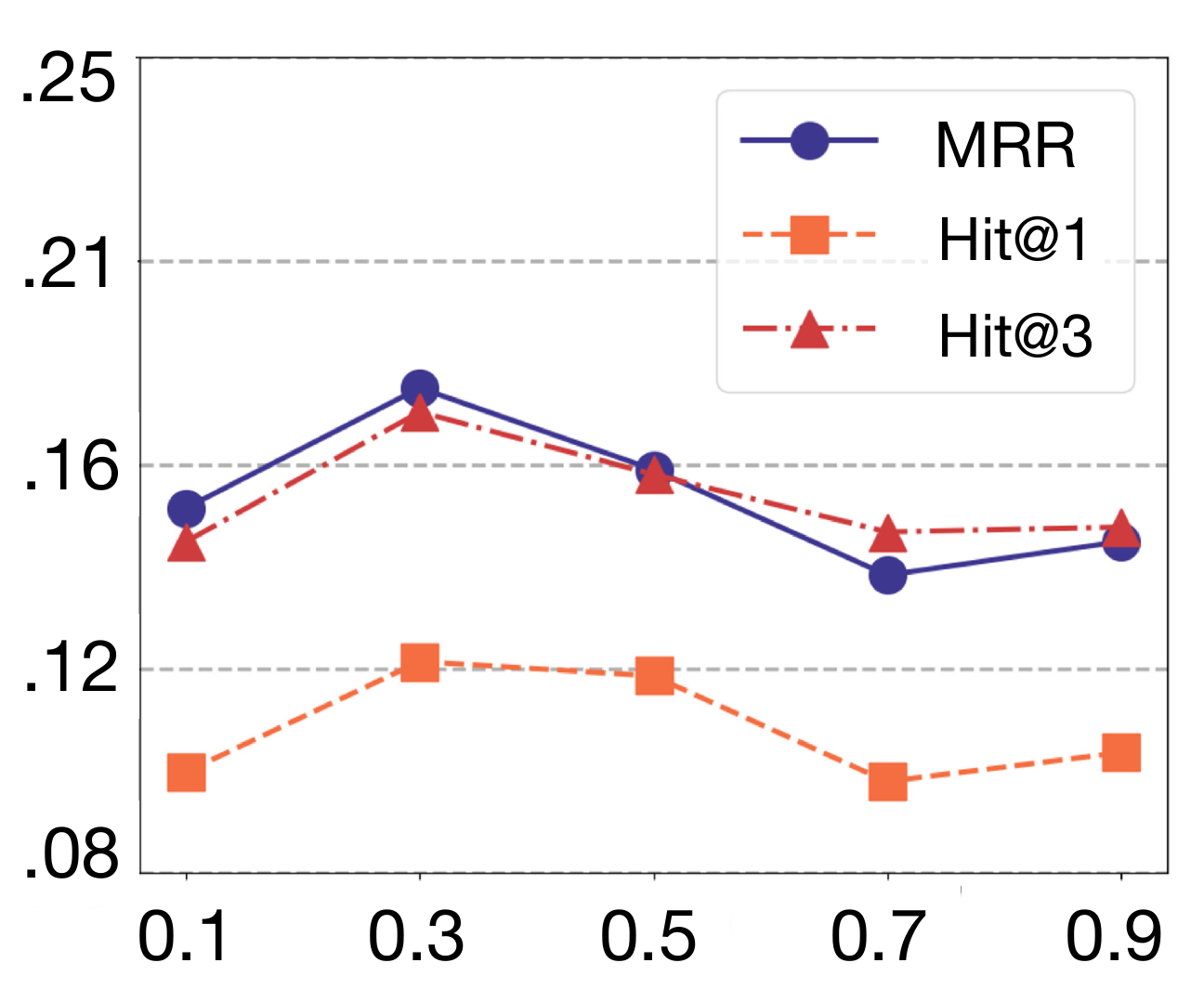}\label{fig:w_2 plot}}
  \caption{Effect of expert weights.}
  \label{fig:weight of experts}
\end{figure}

The results show that model performance peaks when the KB Expert weight \( w_1 \) is around 0.7 and the Hyper-Enz weight \( w_2 \) is about 0.3. This supports our ablation findings: removing the KB Expert degrades performance more than removing the Hyper-Enz Expert. Although \( w_2 \) is smaller, it still contributes positively. The reason may be that Hyper-Enz complements the KB Expert by capturing domain signals from incomplete reactions that the KB Expert might otherwise miss. We conduct a further horizontal comparison experiment among experts in Appendix \ref{app_horizontal}.

\subsection{Case Study}

To qualitatively assess the performance of our method, we present a case study involving a test instance from the dataset. For a testing enzyme-substrate pair: \{Substrate: \textit{D-psicose},  Enzyme: \textit{Q6QWR1}\}, there is a corresponding incomplete biochemical reaction,
\begin{center}
\texttt{D-psicose $\rightarrow$ D-allose}
\end{center}
. We compare the predictions made by several models and experts, as shown in Table~\ref{tab:case-study}. The rank of the ground-truth enzyme as well as the top-3 predicted candidates are reported for each method.

\begin{table}[h]
\centering
\caption{Model predictions and expert responses.}
\label{tab:case-study}
\begin{tabular}{lcc}
\toprule
\textbf{Method} & \textbf{Rank of GT} & \textbf{Top-3 Predictions} \\
\midrule
Boost-RS       & 628 & Q8TGI8, \ Q5BM20,\ P00791 \\
KB-Exp    & --  & - \\
ML-Exp     & 205 & Q52383, \ Q9M9F5, \ P19971 \\
Hyper-Enz & \textbf{1} & \textbf{Q6QWR1}, \ Q04631, \ Q8TGI8 \\
\bottomrule
\end{tabular}
\end{table}


These results demonstrate the strength of our hypergraph-based learning framework. Even when the explicit knowledge base contains incomplete information, our model can leverage collaboration signals to complete reactions and guide accurate enzyme predictions. Further practical impacts are illustrated in Appendix \ref{app_impact}.

\section{Related Work}
\paragraph{\bf Enzyme–substrate interaction prediction.} ML-based enzyme-substrate pair prediction remains under-explored, with most existing methods relying on traditional machine learning. These approaches train directly on enzyme-substrate pairs, incorporating chemical features as auxiliary information. Boost-RS \cite{2022_boost_rs} introduces a framework that enhances traditional collaborative filtering with auxiliary tasks like molecular fingerprint and compound–compound bio-transformation prediction. Enz-Rank \cite{enz_rank2023} employs a CNN-based model using enzyme sequences and substrate molecular signatures to compute probability scores. To improve accuracy, ESP \cite{2023_nature_comm} shifts focus to substrate prediction, using a transformer \cite{Rives622803} and GNN-generated molecular fingerprints with augmented negative examples. ReactZyme \cite{hua2024reactzyme} provides a benchmark for enzyme-reaction prediction, integrating structural and sequence-based features into a standardized dataset for evaluating models leveraging molecular conformations and protein structures.

Note that ReactZyme and our method both frame enzyme-reaction prediction as a retrieval task. While ReactZyme learns a similarity matrix using pretrained molecule and protein encoders, our approach focuses on capturing collaborative patterns among reactions, especially from incomplete equations. Detailed comparison with ReactZyme is provided in Appendix~\ref{appendix:reactzyme}.

\paragraph{\bf Knowledge graph embedding.}
KGE aims to map the entities and relations in KGs to low-dimensional vectors and capture their semantics. KGE models measure the confidence of a triple $(h,r,t)$ using a scoring function $f_r(h,t)$. 
Among them, TransE \cite{transe}, RotatE \cite{rotate}, Rot-Pro \cite{rot-pro} and PairRE \cite{pairre}, HousE \cite{house} are representative of designing the distance-based $f_r$, which assume that after a relation-specific geometric transformation, the position of $h$ in $n$ dimension space is close to $t$. DistMult \cite{distmult}, ComplEx \cite{CompLex}, TuckER \cite{TuckER}, etc. design $f_r$ based on the similarity among the embedding of $h,r$ and $t$. Deep learning models such as ConvE \cite{conve} and ConvKB \cite{convkb} apply convolutional neural networks to learning deep expressive features, and with the popularity of graph neural networks (GNNs), GNN-based models like GAATs \cite{gaat} and NBFNet \cite{zhu2021neural} have been introduced to learn the knowledge representation.

\paragraph{\bf Hypergraph Learning.} 

Hypergraphs inherently represent and capture diverse high-order structures, a technique frequently employed in recommendation systems. Initial works, such as HGNN \cite{aaai2018HGNN} and HyperGCN \cite{nips2019HGCN}, pioneered applying graph convolution to hypergraphs, laying the foundation for this field. However, these methods faced computational and memory inefficiencies due to their reliance on conventional graph convolution. To address this, AllSet \cite{chien2021you} introduced a novel message-passing framework combining Deep Sets and Set Transformers, enhancing modeling flexibility and efficiency, making hypergraph convolution more scalable. Building on these foundations, HCCF \cite{sigir2022HCCF} extended hypergraph convolution into collaborative filtering by capturing both local and global collaborative relations. STHGCN \cite{sthgcn} utilizes hypergraph convolution networks to model diverse user behaviors in Next-POI recommendation, introducing hyperedge collaboration among trajectory graphs.

\section{Conclusion and Future Work}
In this paper, we explored leveraging domain knowledge to improve enzyme prediction accuracy in information retrieval tasks. We introduced a hypergraph-enhanced KGE model, treating equations as triples in the KG. By learning relationships between educts and products with respect to enzymes from both complete and incomplete equations, we complete missing enzyme data within the domain knowledge. Based on this, we proposed an equation-level enzyme prediction task and curated a corresponding dataset. We also introduced a multi-expert prediction paradigm, providing results from three perspectives: enzyme-compound pairs, equations with enzyme information, and those without. Integrating these outputs yields the final prediction. Results show our Hyper-Enz model significantly outperforms baselines at both equation and pair levels. In future work, we would expand equation-level annotations and consider causal inference methods to better utilize reaction data for guiding enzyme–substrate prediction.

\bibliographystyle{ACM-Reference-Format}
\bibliography{ref}

@inproceedings{caisimple,
  title={LightGCL: Simple Yet Effective Graph Contrastive Learning for Recommendation},
  author={Cai, Xuheng and Huang, Chao and Xia, Lianghao and Ren, Xubin},
  booktitle={The Eleventh International Conference on Learning Representations},
  year={2023}
}

@inproceedings{zhang2024heuristic,
  title={Heuristic Learning with Graph Neural Networks: A Unified Framework for Link Prediction},
  author={Zhang, Juzheng and Wei, Lanning and Xu, Zhen and Yao, Quanming},
  booktitle={Proceedings of the 30th ACM SIGKDD Conference on Knowledge Discovery and Data Mining},
  year={2024}
}

@article{FusionESP,
author = {Du, Zhenjiao and Fu, Weimin and Guo, Xiaolong and Caragea, Doina and Li, Yonghui},
title = {FusionESP: Improved Enzyme–Substrate Pair Prediction by Fusing Protein and Chemical Knowledge},
journal = {Journal of Chemical Information and Modeling},
volume = {65},
number = {6},
pages = {2806-2817},
year = {2025},
doi = {10.1021/acs.jcim.4c02357},
    note ={PMID: 40035691},

}

@article{mei,
author = {Qian, Wenjia and Wang, Xiaorui and Huang, Yuansheng and Kang, Yu and Pan, Peichen and Hsieh, Chang-Yu and Hou, Tingjun},
title = {Deep Learning-Driven Insights into Enzyme–Substrate Interaction Discovery},
journal = {Journal of Chemical Information and Modeling},
volume = {65},
number = {1},
pages = {187-200},
year = {2025},
doi = {10.1021/acs.jcim.4c01801},
    note ={PMID: 39721977},
}

@inproceedings{transe,
title = {Translating Embeddings for Modeling Multi-relational Data},
author = {Bordes, Antoine and Usunier, Nicolas and Garcia-Duran, Alberto and Weston, Jason and Yakhnenko, Oksana},
booktitle = {Advances in Neural Information Processing Systems 26},
pages = {2787--2795},
year = {2013},
}

@ARTICLE{gaat,  
author={R. {Wang} and B. {Li} and S. {Hu} and W. {Du} and M. {Zhang}},  journal={IEEE Access},  title={Knowledge Graph Embedding via Graph Attenuated Attention Networks},   
year={2020},  
volume={8}, 
pages={5212--5224}
}

@article{conve,
  title={Convolutional 2D
Knowledge Graph Embeddings},
  author={Tim Dettmers and Pasquale Minervini and Pontus Stenetorp and Sebastian Riedel},
  journal={Proceedings of the 32nd AAAI Conference on Artificial Intelligence},
  year={2018},
}

@inproceedings{convkb,
author = {Dai Quoc Nguyen and Tu Dinh Nguyen and Dat Quoc
Nguyen and Dinh Phung},
year = {2018},
pages = {327–333},
title = {A Novel Embedding Model for Knowledge Base Completion Based on Convolutional Neural Network},
booktitle = {Proceedings of the
2018 Conference of the North American Chapter of
the Association for Computational Linguistics: Human Language Technologies (NAACL)},
volume = {2}
}

@inproceedings{CompLex,
author={T. Trouillon and J. Welbl and S. Riedel and E. Gaussier and G. Bouchard},
year={2016},
booktitle = {Proceedings of 33rd Int. Conf. Mach. Learn},
title = {Complex embeddings for simple link prediction},
pages={2071–2080}
}

@inproceedings{distmult,
author={B. Yang and W.-t. Yih and X. He and J. Gao and L. Deng},
year={2015},
booktitle = {ICLR},
title = {Embedding entities and relations for learning and inference in knowledge bases},
pages={1-13}
}

@inproceedings{GCN,
author = {Kipf, Thomas and Welling, Max},
year = {2017},
pages = {1-14},
title = {Semi-Supervised Classification with Graph Convolutional Networks},
booktitle = {Proceedings of ICLR}
}

@inproceedings{rotate,
 title={RotatE: Knowledge Graph Embedding by Relational Rotation in Complex Space},
 author={Zhiqing Sun and Zhi-Hong Deng and Jian-Yun Nie and Jian Tang},
 booktitle={International Conference on Learning Representations},
 year={2019},
}

@inproceedings{rot-pro,
  title={Rot-Pro: Modeling Transitivity by Projection in Knowledge Graph Embedding},
  author = {Tengwei Song and Jie Luo and Lei Huang},
  booktitle={Proceedings of the Thirty-Fifth Annual Conference on Advances in Neural Information Processing Systems ({NeurIPS})},
  year={2021}
}

@inproceedings{pairre,
    title = "{P}air{RE}: Knowledge Graph Embeddings via Paired Relation Vectors",
    author = "Chao, Linlin  and
      He, Jianshan  and
      Wang, Taifeng  and
      Chu, Wei",
    booktitle = "Proceedings of the 59th Annual Meeting of the Association for Computational Linguistics and the 11th International Joint Conference on Natural Language Processing (Volume 1: Long Papers)",
    month = aug,
    year = "2021",
    address = "Online",
    publisher = "Association for Computational Linguistics",
    pages = "4360--4369"
}

@article{TuckER,
   title={TuckER: Tensor Factorization for Knowledge Graph Completion},
  author={Bala{\v{z}}evi{\'c}, Ivana and Allen, Carl and Hospedales, Timothy},
  booktitle={Proceedings of the 2019 Conference on Empirical Methods in Natural Language Processing and the 9th International Joint Conference on Natural Language Processing (EMNLP-IJCNLP)},
  pages={5185-5194},
  year={2019}
}

@article{aaai2018HGNN,
  title={Hypergraph Neural Networks},
  author={Feng, Yifan and You, Haoxuan and Zhang, Zizhao and Ji, Rongrong and Gao, Yue},
  journal={AAAI 2019},
  year={2018}
}

@inbook{nips2019HGCN,
author = {Yadati, Naganand and Nimishakavi, Madhav and Yadav, Prateek and Nitin, Vikram and Louis, Anand and Talukdar, Partha},
title = {HyperGCN: A New Method of Training Graph Convolutional Networks on Hypergraphs},
year = {2019},
booktitle = {Proceedings of the 33rd International Conference on Neural Information Processing Systems},
articleno = {135},
numpages = {12}
}

@article{chien2021you,
  title={You are allset: A multiset function framework for hypergraph neural networks},
  author={Chien, Eli and Pan, Chao and Peng, Jianhao and Milenkovic, Olgica},
  journal={arXiv preprint arXiv:2106.13264},
  year={2021}
}

@inproceedings{sigir2022HCCF,
author = {Xia, Lianghao and Huang, Chao and Xu, Yong and Zhao, Jiashu and Yin, Dawei and Huang, Jimmy},
title = {Hypergraph Contrastive Collaborative Filtering},
year = {2022},
isbn = {9781450387323},
booktitle = {Proceedings of the 45th International ACM SIGIR Conference on Research and Development in Information Retrieval},
pages = {70–79},
numpages = {10},
series = {SIGIR '22}
}

@InProceedings{house,
  title = 	 {{H}ous{E}: Knowledge Graph Embedding with Householder Parameterization},
  author =       {Li, Rui and Zhao, Jianan and Li, Chaozhuo and He, Di and Wang, Yiqi and Liu, Yuming and Sun, Hao and Wang, Senzhang and Deng, Weiwei and Shen, Yanming and Xie, Xing and Zhang, Qi},
  booktitle = 	 {Proceedings of the 39th International Conference on Machine Learning},
  pages = 	 {13209--13224},
  year = 	 {2022},
  editor = 	 {Chaudhuri, Kamalika and Jegelka, Stefanie and Song, Le and Szepesvari, Csaba and Niu, Gang and Sabato, Sivan},
  volume = 	 {162},
  series = 	 {Proceedings of Machine Learning Research},
  month = 	 {17--23 Jul},
  publisher =    {PMLR}
}

@article{2022_boost_rs,
    author = {Li, Xinmeng and Liu, Li-Ping and Hassoun, Soha},
    title = "{Boost-RS: boosted embeddings for recommender systems and its application to enzyme–substrate interaction prediction}",
    journal = {Bioinformatics},
    volume = {38},
    number = {10},
    pages = {2832-2838},
    year = {2022},
    month = {04},
    issn = {1367-4803},
    doi = {10.1093/bioinformatics/btac201},
  
}

@article{2023_nature_comm,
    author = {Kroll, Alexander and Ranjan, Sahasra and Engqvist, Martin K. M. and Lercher, Martin J.},
    title = "{A general model to predict small molecule substrates of enzymes based on machine and deep learning.}",
    journal = {Nature Communications},
    volume = {14},
    number = {1},
    pages = {2787},
    year = {2023},
    month = {03},
    issn = {2041-1723},
    doi = {10.1038/s41467-023-38347-2},
}

@article{enz_rank2023,
title = {Rank-ordering of known enzymes as starting points for re-engineering novel substrate activity using a convolutional neural network},
journal = {Metabolic Engineering},
volume = {78},
pages = {171-182},
year = {2023},
issn = {1096-7176},
author = {Vikas Upadhyay and Veda Sheersh Boorla and Costas D. Maranas}
}

@article{BRENDA,
title = {The BRENDA enzyme information system–From a database to an expert system},
journal = {Journal of Biotechnology},
volume = {261},
pages = {194-206},
year = {2017},
note = {Bioinformatics Solutions for Big Data Analysis in Life Sciences presented by the German Network for Bioinformatics Infrastructure},
issn = {0168-1656},
doi = {https://doi.org/10.1016/j.jbiotec.2017.04.020},
author = {I. Schomburg and L. Jeske and M. Ulbrich and S. Placzek and A. Chang and D. Schomburg}
}

@article{kegg,
  title={KEGG: new perspectives on genomes, pathways, diseases and drugs},
  author={Kanehisa, Minoru and Furumichi, Miho and Tanabe, Mao and Sato, Yoko and Morishima, Kanae},
  journal={Nucleic Acids Research},
  volume={45},
  number={D1},
  pages={D353--D361},
  year={2017},
  publisher={Oxford University Press},
  doi={10.1093/nar/gkw1092},
  url={http://www.kegg.jp/ or http://www.genome.jp/kegg/},
  pmid={27899662},
  pmc={PMC5210567},
  note={Epub 2016 Nov 28}
}

@article{pubchem2023,
    author = {Kim, Sunghwan and Chen, Jie and Cheng, Tiejun and Gindulyte, Asta and He, Jia and He, Siqian and Li, Qingliang and Shoemaker, Benjamin A and Thiessen, Paul A and Yu, Bo and Zaslavsky, Leonid and Zhang, Jian and Bolton, Evan E},
    title = "{PubChem 2023 update}",
    journal = {Nucleic Acids Research},
    volume = {51},
    number = {D1},
    pages = {D1373-D1380},
    year = {2022},
    month = {10},
    issn = {0305-1048},
    doi = {10.1093/nar/gkac956},
    url = {https://doi.org/10.1093/nar/gkac956},
    eprint = {https://academic.oup.com/nar/article-pdf/51/D1/D1373/48441598/gkac956.pdf},
}

@article{BKM_react,
author = {Lang Maren and Stelzer Michael and Schomburg Dietmar},
year = {2011},
month = {08},
pages = {42},
title = {BKM-react, an integrated biochemical reaction database},
volume = {12},
journal = {BMC biochemistry},
doi = {10.1186/1471-2091-12-42}
}

@article{Askr2023,
  author    = {Heba Askr and Enas Elgeldawi and Heba Aboul Ella and Yaseen A. M. M. Elshaier and Mamdouh M. Gomaa and Aboul Ella Hassanien},
  title     = {Deep learning in drug discovery: an integrative review and future challenges},
  journal   = {Artificial Intelligence Review},
  year      = {2023},
  volume    = {56},
  number    = {7},
  pages     = {5975--6037},
  doi       = {10.1007/s10462-022-10306-1},
  url       = {https://doi.org/10.1007/s10462-022-10306-1},
  issn      = {1573-7462}
}

@article{biofuels,
title = {Biofuels and their sources of production: A review on cleaner sustainable alternative against conventional fuel, in the framework of the food and energy nexus},
journal = {Energy Nexus},
volume = {4},
pages = {100036},
year = {2021},
issn = {2772-4271},
doi = {https://doi.org/10.1016/j.nexus.2021.100036},
url = {https://www.sciencedirect.com/science/article/pii/S277242712100036X},
author = {Sangita Mahapatra and Dilip Kumar and Brajesh Singh and Pravin Kumar Sachan},
}

@article {Rives622803,
	author = {Rives, Alexander and Meier, Joshua and Sercu, Tom and Goyal, Siddharth and Lin, Zeming and Liu, Jason and Guo, Demi and Ott, Myle and Zitnick, C. Lawrence and Ma, Jerry and Fergus, Rob},
	title = {Biological structure and function emerge from scaling unsupervised learning to 250 million protein sequences},
	elocation-id = {622803},
	year = {2020},
	doi = {10.1101/622803},
	publisher = {Cold Spring Harbor Laboratory},
	URL = {https://www.biorxiv.org/content/early/2020/12/15/622803},
	eprint = {https://www.biorxiv.org/content/early/2020/12/15/622803.full.pdf},
	journal = {bioRxiv}
}

@inproceedings{iclr2021ALLSET,
author = {Chien, Eli and Pan, Chao and Peng, Jianhao and Milenkovic, Olgica},
year = {2021},
month = {06},
pages = {},
title = {You are AllSet: A Multiset Function Framework for Hypergraph Neural Networks}
}

@inproceedings{sthgcn,
author = {Yan, Xiaodong and Song, Tengwei and Jiao, Yifeng and He, Jianshan and Wang, Jiaotuan and Li, Ruopeng and Chu, Wei},
title = {Spatio-Temporal Hypergraph Learning for Next POI Recommendation},
year = {2023},
isbn = {9781450394086},
publisher = {Association for Computing Machinery},
address = {New York, NY, USA},
booktitle = {Proceedings of the 46th International ACM SIGIR Conference on Research and Development in Information Retrieval},
pages = {403–412},
numpages = {10},
location = {Taipei, Taiwan},
series = {SIGIR '23}
}

@article{zhu2021neural,
  title={Neural bellman-ford networks: A general graph neural network framework for link prediction},
  author={Zhu, Zhaocheng and Zhang, Zuobai and Xhonneux, Louis-Pascal and Tang, Jian},
  journal={Advances in Neural Information Processing Systems},
  volume={34},
  year={2021}
}

@inproceedings{pytorch,
 author = {Paszke, Adam and Gross, Sam and Massa, Francisco and Lerer, Adam and Bradbury, James and Chanan, Gregory and Killeen, Trevor and Lin, Zeming and Gimelshein, Natalia and Antiga, Luca and Desmaison, Alban and Kopf, Andreas and Yang, Edward and DeVito, Zachary and Raison, Martin and Tejani, Alykhan and Chilamkurthy, Sasank and Steiner, Benoit and Fang, Lu and Bai, Junjie and Chintala, Soumith},
 booktitle = {Advances in Neural Information Processing Systems},
 editor = {H. Wallach and H. Larochelle and A. Beygelzimer and F. d\textquotesingle Alch\'{e}-Buc and E. Fox and R. Garnett},
 pages = {},
 publisher = {Curran Associates, Inc.},
 title = {PyTorch: An Imperative Style, High-Performance Deep Learning Library},
 url = {https://proceedings.neurips.cc/paper_files/paper/2019/file/bdbca288fee7f92f2bfa9f7012727740-Paper.pdf},
 volume = {32},
 year = {2019}
}

@inproceedings{
shazeer2017,
title={ Outrageously Large Neural Networks: The Sparsely-Gated Mixture-of-Experts Layer},
author={Noam Shazeer and *Azalia Mirhoseini and *Krzysztof Maziarz and Andy Davis and Quoc Le and Geoffrey Hinton and Jeff Dean},
booktitle={International Conference on Learning Representations},
year={2017},
url={https://openreview.net/forum?id=B1ckMDqlg}
}

@inproceedings{DMF,
author = {Xue, Hong-Jian and Dai, Xin-Yu and Zhang, Jianbing and Huang, Shujian and Chen, Jiajun},
title = {Deep matrix factorization models for recommender systems},
year = {2017},
isbn = {9780999241103},
publisher = {AAAI Press},
booktitle = {Proceedings of the 26th International Joint Conference on Artificial Intelligence},
pages = {3203–3209},
numpages = {7},
location = {Melbourne, Australia},
series = {IJCAI'17}
}

@inproceedings{NGCF,
author = {Wang, Xiang and He, Xiangnan and Wang, Meng and Feng, Fuli and Chua, Tat-Seng},
title = {Neural Graph Collaborative Filtering},
year = {2019},
isbn = {9781450361729},
publisher = {Association for Computing Machinery},
address = {New York, NY, USA},
url = {https://doi.org/10.1145/3331184.3331267},
doi = {10.1145/3331184.3331267},
booktitle = {Proceedings of the 42nd International ACM SIGIR Conference on Research and Development in Information Retrieval},
pages = {165–174},
numpages = {10},
location = {Paris, France},
series = {SIGIR'19}
}

@article{NCF,
  author       = {Xiangnan He and
                  Lizi Liao and
                  Hanwang Zhang and
                  Liqiang Nie and
                  Xia Hu and
                  Tat{-}Seng Chua},
  title        = {Neural Collaborative Filtering},
  journal      = {CoRR},
  volume       = {abs/1708.05031},
  year         = {2017},
  eprinttype    = {arXiv},
  eprint       = {1708.05031}
}

@inproceedings{bpr,
author = {Rendle, Steffen and Freudenthaler, Christoph and Gantner, Zeno and Schmidt-Thieme, Lars},
title = {BPR: Bayesian personalized ranking from implicit feedback},
year = {2009},
isbn = {9780974903958},
publisher = {AUAI Press},
address = {Arlington, Virginia, USA},
booktitle = {Proceedings of the Twenty-Fifth Conference on Uncertainty in Artificial Intelligence},
pages = {452–461},
numpages = {10},
location = {Montreal, Quebec, Canada},
series = {UAI '09}
}

@inproceedings{hua2024reactzyme,
title={ReactZyme: A Benchmark for Enzyme-Reaction Prediction},
author={Chenqing Hua and Bozitao Zhong and Sitao Luan and Liang Hong and Guy Wolf and Doina Precup and Shuangjia Zheng},
booktitle={The Thirty-eight Conference on Neural Information Processing Systems Datasets and Benchmarks Track},
year={2024},
url={https://openreview.net/forum?id=xepxnDQoGq}
}

@inproceedings{DGCF,
author = {Wang, Xiang and Jin, Hongye and Zhang, An and He, Xiangnan and Xu, Tong and Chua, Tat-Seng},
title = {Disentangled Graph Collaborative Filtering},
year = {2020},
isbn = {9781450380164},
publisher = {Association for Computing Machinery},
address = {New York, NY, USA},
url = {https://doi.org/10.1145/3397271.3401137},
doi = {10.1145/3397271.3401137},
booktitle = {Proceedings of the 43rd International ACM SIGIR Conference on Research and Development in Information Retrieval},
pages = {1001–1010},
numpages = {10},
location = {Virtual Event, China},
series = {SIGIR '20}
}

\appendix
\section{Relation Patterns in Equation KGs}
\label{sec:relation-patterns}

Equation knowledge graphs differ from traditional KGs in that their relations (enzymes) represent biochemical transformations, often exhibiting unique relational patterns rooted in reaction reversibility and enzyme-specificity. To accurately model such reaction-level semantics, a KGE model must be capable of capturing various relation patterns. In this work, we focus on three fundamental patterns observed in equation KGs:

\begin{itemize}
    \item \textbf{Symmetry}: A relation $m$ is symmetric if $\langle S, m, P \rangle$ implies $\langle P, m, S \rangle$. This corresponds to a reversible reaction catalyzed by the same enzyme.
    \item \textbf{Asymmetry}: A relation $m$ is asymmetric if $\langle S, m, P \rangle$ does not imply $\langle P, m, S \rangle$. This reflects an irreversible reaction.
    \item \textbf{Inversion}: Two relations $m_1$ and $m_2$ are inverses if $\langle S, m_1, P \rangle$ implies $\langle P, m_2, S \rangle$. This corresponds to reversibility catalyzed by different enzymes.
\end{itemize}

To support all these patterns, we adopt \textbf{PairRE}~\cite{pairre}, a distance-based embedding model which generalizes TransE by allowing each relation to learn a pair of vectors for head and tail transformations. Specifically, for a triple $(h, r, t)$, PairRE defines the scoring function as:

\[
f(h, r, t) = -\left\| \mathbf{h} \circ \mathbf{r}_h - \mathbf{t} \circ \mathbf{r}_t \right\|_p
\]

where $\mathbf{h}, \mathbf{t} \in \mathbb{R}^d$ are entity embeddings, $\mathbf{r}_h, \mathbf{r}_t \in \mathbb{R}^d$ are relation-specific vectors for head and tail transformations, and $\circ$ denotes the Hadamard (element-wise) product.

\paragraph{\bf Capability of Modeling Relation Patterns}
We now show that PairRE can represent the three key patterns:

\paragraph{Symmetry.}  
Suppose $\mathbf{r}_h = \mathbf{r}_t = \mathbf{r}$. Then for a true triple $(h, r, t)$, the model enforces:

\[
\mathbf{h} \circ \mathbf{r} \approx \mathbf{t} \circ \mathbf{r}
\Rightarrow \mathbf{t} \circ \mathbf{r} \approx \mathbf{h} \circ \mathbf{r}
\]

which implies $(t, r, h)$ is also a valid triple, hence symmetry is satisfied.

\paragraph{Asymmetry.}  
Let $\mathbf{r}_h \ne \mathbf{r}_t$ and not constrained to be symmetric. Then $(h, r, t)$ having low distance does not imply $(t, r, h)$ will also have low distance, since:

\[
\mathbf{t} \circ \mathbf{r}_h \not\approx \mathbf{h} \circ \mathbf{r}_t
\]

Thus, asymmetry can be modeled.

\paragraph{Inverse.}  
Let $r_1$ and $r_2$ be inverse relations. PairRE can set:

\[
\mathbf{r}_{1h} = \mathbf{r}_{2t}, \quad \mathbf{r}_{1t} = \mathbf{r}_{2h}
\]

Then for $(h, r_1, t)$, we have:

\[
\mathbf{h} \circ \mathbf{r}_{1h} \approx \mathbf{t} \circ \mathbf{r}_{1t}
\Rightarrow \mathbf{t} \circ \mathbf{r}_{2h} \approx \mathbf{h} \circ \mathbf{r}_{2t}
\]

which makes $(t, r_2, h)$ a valid triple, satisfying the inverse pattern.

\section{Comparison with ReactZyme}
\label{appendix:reactzyme}

\paragraph{\bf Model difference.}Both ReactZyme and our method formulate enzyme-reaction prediction as a retrieval task—ranking candidate enzymes by their catalytic relevance to a given reaction. However, ReactZyme relies on structured biological input, which uses SMILES strings for compounds and protein sequences for enzymes, which are encoded by pre-trained molecular and protein encoders (e.g., ESM and AlphaFold). The model computes reaction embeddings via cross-attention between substrates and products, and retrieves enzymes based on learned similarity in the embedding space.

In contrast, our method does not use explicit chemical or protein features. Instead, it treats compounds and enzymes as symbolic entities with randomly initialized embeddings, and learns their representations purely from the statistical and structural co-occurrence across reaction equations. This abstraction allows the model to focus on relational patterns within the reaction network, rather than relying on specific domain encoders.

\paragraph{\bf Dataset comparison and alignment}
Table \ref{tab:app_dataset_comp} is the dataset comparison of our dataset with ReactZyme, where we use the definition of 
$RelationDensity = \frac{| T |}{|R |}$, and $EntityDensity = \frac{|2T|}{|E|}$, which measures the data distribution of a KG. Correspondingly, the $RelationDensity$ refers to the density of enzymes and the $EntityDensity$ refers to the density of compounds.

\begin{table}[h]
\centering
\caption{Dataset Comparison}
\label{tab:app_dataset_comp}
\begin{tabular}{lccccc}
\toprule
 & \multicolumn{2}{c}{Compounds} & \multicolumn{2}{c}{Enzymes} & \\
\cline{2-5}
 & Count & Density & Count & Density & \\
\midrule
ReactZyme & 400,316 & 83.80 & 9,554 & 2.2 & \\
Ours & 53,399 & 1.36 & 39,044 & 2.5 & \\
Ours-Align & 8,012 & 4.18 & 8,506 & 2.0 \\
\bottomrule
\end{tabular}
\end{table}

It first can be observed that the compound density in ReactZyme is significantly higher than in our dataset. Secondly, to equip ReactZyme with essential features, we further aligned enzymes and compounds in our equation dataset. Due to missing SMILES strings for some compounds, the dataset used in this experiment is a small subset of the full knowledge graph, as shown by Ours-Align in the Table \ref{tab:app_dataset_comp}.


\begin{table}[h]
\centering
\caption{Performance comparison on enzyme ranking with ReactZyme baselines.}
\label{tab:reactzyme-comparison}
\begin{tabular}{lccccc}
\toprule
Model & MRR & MR & Hit@1 & Hit@3 & Hit@10 \\
\midrule
Ours & \textbf{0.151} & 2057 & \textbf{0.088} & \textbf{0.167} & \textbf{0.275} \\
ReactZyme-MLP & 0.043 & 401 & 0.009 & 0.028 & 0.097 \\
ReactZyme-RNN & 0.047 & 360 & 0.010 & 0.028 & 0.099 \\
\bottomrule
\end{tabular}
\end{table}

\paragraph{\bf Alignment results.} We report standard retrieval metrics including Mean Reciprocal Rank (MRR), Mean Rank (MR), Hit@1, Hit@3, and Hit@10. As shown in Table~\ref{tab:reactzyme-comparison}, ReactZyme performs relatively poorly overall, possibly due to the limited dataset size. However, it achieves a notably lower MR, which may stem from its use of pretrained encoders that align compound and protein representations into a unified embedding space. In contrast, our method starts from randomly initialized embeddings, which can lead to insufficient training under sparse data conditions. Nevertheless, our model still significantly outperforms both ReactZyme-MLP and ReactZyme-RNN across most evaluation metrics. This performance gap highlights the advantage of explicitly modeling collaborative information in our framework.

\section{Horizontal comparison among experts.}

To further analyze the contribution of each expert module, we conducted an ablation study and weight sensitivity analysis among the three experts. We first divided the ES-23k dataset into four mutually exclusive subsets based on the completeness of reaction information associated with each target substrate: 
\begin{enumerate}
    \item $S_1$, where the substrate matches only complete reactions; 
    \item $S_2$, where it matches only incomplete reactions; 
    \item $S_3$, where both types are matched; 
    \item $S_4$, where no matching reaction exists.
\end{enumerate}

\begin{table}[h]
\centering
\caption{Performance of different expert combinations on stratified subsets of the ES-23k dataset.}
\begin{tabular}{lcccc}
\toprule
\textbf{Expert Setting} & \textbf{$S_1$} & \textbf{$S_2$} & \textbf{$S_3$} & \textbf{$S_4$} \\
\midrule
ML Expert                 & 0.32     & 0.28     & 0.33     & 0.32 \\
KB + ML Expert               & 0.62   & 0.28     & 0.42   & 0.32   \\
Hyper-Enz + ML Expert        & 0.47     & 0.45   & 0.45   & 0.32   \\
KB + Hyper-Enz + ML Expert   & 0.60    & 0.45     & 0.48   & 0.32   \\
\bottomrule
\end{tabular}
\label{tab:horizontal_comparison}
\end{table}

On these subsets, we evaluated several expert configurations, including the standalone ML expert, the combination of KB and ML experts, Hyper-Enz and ML experts, and the full integration of KB, Hyper-Enz, and ML experts. The results are shown in Table~\ref{tab:horizontal_comparison}. 

Here are some key findings from the horizontal comparison experiment:
\begin{itemize}
    \item ML expert gains Hit@10 = 0.32 in $S_4$, where no equation data exists, illustrating the base capability of the ML expert.
    \item KB + ML expert gains 0.62 on $S_1$ and 0.42 on $S_3$, showing the benefits of leveraging domain knowledge to improve performance when complete reaction data is present.
    \item Hyper-Enz + ML expert gains results around 0.45 on $S_{1,2,3}$, showing that the base capability of Hyper-Enz from complete or partial reactions can effectively enhance learning.
    \item The full combination of KB, Hyper-Enz, and ML experts on $S_1$ shows a slightly decrease compared to KB + ML expert, showing that Hyper-Enz would introduce some noise to KB expert when the equations are complete.  However, it leads to better performance on mixed-reaction subsets, demonstrating its importance in this scenario.
\end{itemize}

\section{Computational Efficiency Analysis}
\label{app_horizontal}
To assess the computational efficiency of our method, we conduct a detailed comparison with representative baselines including TransE, PairRE, and TuckER. We report five metrics: \#Param (total number of parameters, in millions), Train (average training time per epoch, in seconds), Test (average testing time, in seconds), Mem (peak GPU memory usage, in MB), and Complexity (theoretical computational complexity).

\begin{table}[h]
\centering
\caption{Comparison of computational cost between our method and baselines.}
\label{tab:computation-cost}
\begin{tabular}{lccccc}
\toprule
\textbf{Model} & \textbf{\#Param} & \textbf{Train} & \textbf{Test} & \textbf{Mem} & \textbf{Complexity} \\
\midrule
Ours     & 202.08 & 2.90 & 15  & 9182  & $\mathcal{O}(S \cdot d)$ \\
TransE   & 63.27  & 1.12 & 21  & 5452  & $\mathcal{O}(d)$ \\
PairRE   & 96.46  & 1.50 & 28  & 8418  & $\mathcal{O}(d)$ \\
TuckER   & 573.30 & 1.94 & 29  & 8586  & $\mathcal{O}(d^3)$ \\
\bottomrule
\end{tabular}
\end{table}

As shown in Table~\ref{tab:computation-cost}, our model maintains a reasonable trade-off between performance and computational cost. Although it involves a moderately larger number of parameters compared to simpler models like TransE and PairRE, its training time per epoch (2.90s) remains efficient and comparable. Notably, our method achieves a significantly lower testing time (15s), suggesting faster inference capabilities—a crucial factor for large-scale or real-time applications. In terms of GPU memory usage, our model consumes slightly more resources due to its enriched structure but remains within practical bounds. From a complexity perspective, our method exhibits linear complexity with respect to the number of slots $S$ and embedding dimension $d$, denoted as $\mathcal{O}(S \cdot d)$, which is far more scalable than TuckER’s cubic complexity $\mathcal{O}(d^3)$.

Overall, the results demonstrate that our approach strikes a strong balance between effectiveness and efficiency: it incurs only moderate computational cost while outperforming baselines in predictive accuracy, making it a practical choice for real-world deployment.

\subsection{Scalability with Data Scale}

To evaluate the scalability of our framework across different data scales, we conduct computational experiments by varying the number of compounds and enzymes. Table~\ref{tab:scalability} reports four key metrics: \#Param (total parameters in millions), Train (training time per epoch, in seconds), Test (testing time, in seconds), and Memory (GPU memory usage, in MB).

\begin{table}[h]
\centering
\caption{Scalability under varying numbers of compounds and enzymes.}
\label{tab:scalability}
\begin{tabular}{cccccc}
\toprule
\textbf{Comp} & \textbf{Enz} & \textbf{\#Param} & \textbf{Train} & \textbf{Test} & \textbf{Memory } \\
\midrule
39044  & 21153  & 202  & 2.90  & 15  & 9182 \\
15071  & 14607  & 79   & 2.34  & 7   & 8224 \\
13298  & 11176  & 49   & 2.30  & 5   & 8062 \\
\bottomrule
\end{tabular}
\end{table}

As the number of compounds and enzymes decreases, the model shows a clear reduction in computational cost across all metrics, demonstrating its ability to scale efficiently with smaller input graphs. However, when deployed in large biochemical networks, several scalability challenges emerge. 

First, hypergraph construction and node sampling become increasingly expensive as the incidence matrices $H^{(1)}$ and $H^{(2)}$ densify with the growth of compounds and reactions. Towards this issue, our framework adopts the \textbf{fixed neighbor sampling sizes} $(\eta_1, \eta_2)$ to limit computational expansion. Second, the choice of embedding dimension and model depth significantly influences both memory footprint and forward/backward pass complexity.

\section{Practical Impact}
\label{app_impact}
In real-world settings, enzyme-substrate prediction faces several key challenges. By consulting some human domain experts, below, we outline specific examples of the issues that Hyper-Enz can effectively address.
\paragraph{\bf Misleading Co-occurrence in Reaction Equations}
A common strategy to assist enzyme–substrate prediction is to examine biochemical reaction equations and extract co-occurring enzyme–compound pairs.
However, one major practical challenge is that, the presence of a compound in the equation does not necessarily indicate that it is the substrate of the listed enzyme. Instead, it may serve as a co-substrate, cofactor, or byproduct. Blindly assuming that all co-occurring compounds are substrates leads to a high risk of misidentifying enzyme–substrate pairs. Here are some practical examples:

\begin{example}
ATP is commonly seen in kinase-catalyzed reactions, such as the phosphorylation of glucose:

$$
\text{Glucose} + \text{ATP} \xrightarrow{\text{Hexokinase}} \text{Glucose-6-phosphate} + \text{ADP}.
$$

Here, ATP participates in the reaction, but it is not the primary substrate of hexokinase—the enzyme primarily acts on glucose. ATP's role is that of a phosphate donor. In contrast, ATP is the actual substrate of ATPases, which directly hydrolyze ATP. Therefore, while ATP co-occurs in the hexokinase reaction, this co-occurrence does not imply a direct enzyme–substrate relationship between ATP and hexokinase.
\end{example}

\begin{example}
NAD$^+$ appears in dehydrogenase-catalyzed redox reactions, such as:
$$
\text{Lactate} + \text{NAD}^+ \xrightarrow{\text{Lactate dehydrogenase}} \text{Pyruvate} + \text{NADH}.
$$
NAD$^+$ is not the primary substrate of lactate dehydrogenase; lactate is. NAD$^+$ acts as an electron acceptor. However, NAD$^+$ is the actual substrate of enzymes such as NADases or NAD synthases. Thus, NAD$^+$’s presence in this equation does not equate to a substrate role for this enzyme.
\end{example}




Such ambiguity is not rare. In our data, more than 40\% of reported substrates co-occur with five or more enzymes in different reactions, highlighting the risk of spurious pairings. Methods relying solely on equation-level co-occurrence are prone to introducing noise into enzyme–substrate prediction tasks.

\paragraph{\bf Case Study}
We provide a relevant case study of our method for such situation:

For a testing enzyme-substrate pair: \{Substrate: \textit{phenanthridine},  Enzyme: \textit{P48034} (putative)\}, there is a corresponding biochemical reaction,
\begin{center}
\texttt{H$_2$O, O$_2$, phenanthridine$\xrightarrow{\text{P05091}}$6-phenantridone, H$_2$O$_2$}
\end{center}
, found by the KB expert. We compare the predictions made by several models and experts, as shown in Table~\ref{tab:case-study}. The rank of the ground-truth enzyme as well as the top-3 predicted candidates are reported for each method.

\begin{table}[h]
\centering
\caption{Model predictions and expert responses.}
\label{tab:case-study}
\begin{tabular}{lcc}
\toprule
\textbf{Method} & \textbf{Rank of GT} & \textbf{Top-3 Predictions} \\
\midrule
Boost-RS  &  602 & Q52383,\ P21219,\ P00791 \\
KB-Exp    & --  & P05091 \\
ML-Exp     & 194 & P19971,\ Q8TGI8,\ P21219 \\
Hyper-Enz & \textbf{1} & \textbf{P48034},\ B1GV57,\ Q3ZFI \\
\bottomrule
\end{tabular}
\end{table}

We can see that even when the KB Expert providing the misleading information, our Hyper-Enz Expert ranks the correct enzyme \textit{P48034} at the top position.

\paragraph{\bf Expert feedback: Potentials in cascade reactions}

According to human expert feedback, one major challenge in enzyme–substrate prediction lies in accurately resolving cascade (multi-step) reactions, which are commonly seen in metabolic pathways. In such cases, the same compound may act as a product in one enzymatic step and as a substrate in the next, often causing ambiguity in enzyme annotation and prediction. Standard learning approaches that treat substrate–product pairs uniformly may confuse the roles of intermediates, especially when enzymes are missing for intermediate steps or when multiple reactions share overlapping compounds.

\begin{example}
    
For instance, in BRENDA, the following cascade reaction occurs in cobalamin biosynthesis:

\begin{align*}
precorrin\text{-}2&\xrightarrow{CobI\ (Q9X795)} precorrin\text{-}3A \\
precorrin\text{-}3A &\xrightarrow{CobJ\ (Q9X796)} precorrin\text{-}4
\end{align*}

Here, \textit{precorrin-3A} acts both as a product (in the reaction catalyzed by \textit{CobI}) and as a substrate (in the reaction catalyzed by \textit{CobJ}). When the enzyme for the second step (\textit{CobJ}) is missing,  human experts may struggle to determine whether \textit{precorrin-3A} is the intended product or merely an intermediate that undergoes further conversion—especially when multiple pathway variants exist in BRENDA. 

\end{example}

Traditional models relying on co-occurrence or embedding-based similarity often fail to distinguish such directionality. As a result, they may rank unrelated enzymes that act on \textit{precorrin-3A} in different pathways higher than the correct one.
However, our model explicitly distinguishes between substrates and products during hypergraph construction. When the same compound appears as a substrate in one reaction and as a product in another, it is assigned to distinct hyperedges and participates in separate rounds of message passing. This design enables the model to capture the role-specific behavior of compounds and avoid misleading associations caused by role ambiguity. For future research, we believe it would be meaningful to construct a dataset specifically focused on cascading reactions and apply our method to this setting.

\end{document}
\endinput